\documentclass[letterpaper,twocolumn,10pt]{article}
\usepackage{usenix-2020-09}

\usepackage{tikz}
\usepackage{amsmath}
\usepackage{pgfplots} 
\usepgfplotslibrary{statistics}
\usetikzlibrary{patterns}

\newcommand{\eg}{\textit{e.g\@.}}

\newcommand{\ie}{\textit{i.e\@.}}

\newcommand{\mr}[1]{{\color{black} {#1}}}

\newcommand{\rev}[1]{{\color{black} {#1}}}
\usepackage[numbers]{natbib}
\usepackage{amsmath,amssymb,amsfonts}
\usepackage{pdfcomment}
\usepackage{hyperref}
\usepackage{cleveref}
\usepackage{booktabs}
\usepackage{multicol}
\usepackage{multirow}
\usepackage{adjustbox}
\usepackage{tabularx}
\usepackage{subcaption}
\usepackage{adjustbox}
\usepackage{xspace}
\usepackage{microtype}
\usepackage{textcomp, gensymb}
\usepackage{algorithm}
\usepackage{circledsteps}
\usepackage{algorithmic}
\usepackage{amsthm}
\usepackage{enumitem}
\usepackage[font=small]{caption}
\usepackage{authblk}
\makeatletter
\renewcommand\AB@affilsepx{, \protect\Affilfont}
\makeatother

\microtypecontext{spacing=nonfrench}
\pgfplotsset{compat=1.17}

\newcommand{\name}{{\fontfamily{cmss}\selectfont{Mystique}}\xspace}
\newcommand{\ASI}{\textsf{ASI}\xspace}

\usepackage{letltxmacro}
\LetLtxMacro{\originaleqref}{\eqref}
\renewcommand{\eqref}{Eqn.~\originaleqref}

\usepackage{colortbl}
\definecolor{aliceblue}{rgb}{0.94, 0.97, 1.0}

\usepackage{titlesec}

\titlespacing{\paragraph}{%
  0pt}{%
  0.3\baselineskip}{%
  1em}%

\newif\ifpaper
\paperfalse   %

\begin{document}

\date{}

\title{\Large \bf Tubes Among Us: Analog Attack on Automatic Speaker Identification}

\author[1]{\rm Shimaa Ahmed\thanks{Corresponding Author: ahmed27@wisc.edu} }
\author[1]{\rm Yash Wani}
\author[2]{\rm Ali Shahin Shamsabadi\thanks{Work done partially while the author was at the Vector Institute.} }
\author[3,4]{\rm Mohammad Yaghini}
\author[4,5]{\rm  Ilia Shumailov}
\author[3,4]{\\ \rm  Nicolas Papernot}
\author[1]{\rm  Kassem Fawaz}
\affil[1]{\textit{University of Wisconsin-Madison}}
\affil[2]{\textit{Alan Turing Institute}}
\affil[3]{\textit{University of Toronto}}
\affil[4]{\textit{Vector Institute}}
\affil[5]{\textit{University of Oxford}}

\maketitle

\begin{abstract}
    Recent years have seen a surge in the popularity of acoustics-enabled personal devices powered by machine learning. Yet, machine learning has proven to be vulnerable to adversarial examples. A large number of modern systems protect themselves against such attacks by targeting artificiality, \ie, they deploy mechanisms to detect the lack of human involvement in generating the adversarial examples.
    However, these defenses implicitly assume that humans are incapable of producing meaningful and targeted adversarial examples. In this paper, we show that this base assumption is wrong. In particular, we demonstrate that for tasks like speaker identification, a human is capable of producing analog adversarial examples directly with little cost and supervision: by simply speaking through a tube, an adversary reliably impersonates other speakers in eyes of ML models for speaker identification. Our findings extend to a range of other acoustic-biometric tasks such as liveness detection, bringing into question their use in security-critical settings in real life, such as phone banking.
\end{abstract}
\section{Introduction}

As a primary mechanism for human communication, speech is a natural vehicle for human-computer interaction (HCI). Fueled by advancements in Machine Learning (ML), everyday devices and services accept speech as input; users can seamlessly control their smart devices and communicate with automated customer services. This convenience brought the need to authenticate users when speech is the primary interaction modality. Companies deploy \mr{automatic} speaker identification systems (\ASI) that pack ML-based models to authenticate users based on their voiceprint~\cite{Nagrani17,snyder2018}.

\begin{figure}[t]
    \centering
  \includegraphics[width=\columnwidth]{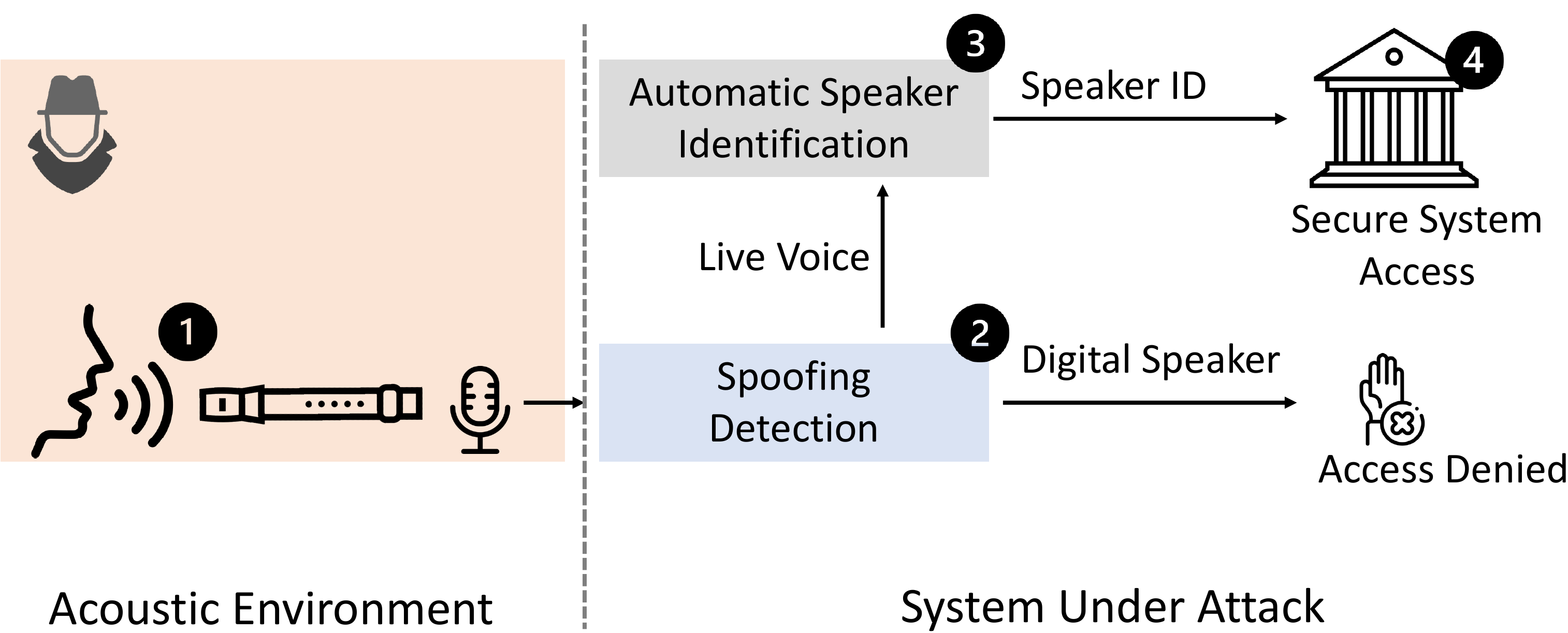}
    \caption{{Overview of \name voice impersonation attack. Left: Acoustic environment fall's under the adversary's control. Right: the system under attack setup. \Circled[fill color=black,inner color=white]{1} The adversary speaks through an adversarially designed  tube. \Circled[fill color=black,inner color=white]{2} A liveness detection model confirms the liveness of the captured voice. \Circled[fill color=black,inner color=white]{3} An automatic speaker identification model recognises the identity of the adversary as the target speaker. \Circled[fill color=black,inner color=white]{4} The secure system gives access to the adversary. }}
    \label{fig:overview}
    \vspace{-5pt}
\end{figure}

Speaker identification systems are vulnerable to an array of attacks such as speech synthesis~\cite{taylor2009text,oord2016wavenet,wang2017tacotron}, voice conversion~\cite{muda2010voice,sisman2018adaptivewavenet,yuan2021improving}, replay attacks~\cite{Lindberg1999vulnerabilityin}, and adversarial examples~\cite{liu2019adversarial,kassis2021practical,chen2021spoofingspeaker}. The adversary generates and feeds the speaker identification system a speech sample to impersonate a target speaker. While the attack techniques differ, they share a common principle: \textit{the attacker manipulates the speech signal in the digital domain} and potentially plays it through a speaker. Note that even physical adversarial examples in the vision domain follow the same principle. Generating these examples requires obtaining a signal (such as a speech recording or a visual patch) by solving an optimization problem in the digital domain and later realizing it in the analog domain.

Current defenses leverage this observation and employ mechanisms to detect the digital attack artifacts in the input signal~\cite{shang2018defending,wang2019defeatinghiddelaudio,wang2019secureyourvoice}. These defenses target either the (1) physical properties of the speaker~\eg~their physical presence~\cite{owczarek2012lipreading,zhang2017hearingyourvoice} or (2) properties of the speech speakers produce~\eg~the energy distribution of different harmonics~\cite{kersta1962voiceprint,chen2017youcanhear}. The resulting unified acoustic pipeline constrains the attacker when generating the attack samples, thus increasing the cost of the attack~\cite{meng2018wivo,shang2018defending,wang2019secureyourvoice}. Generally speaking, the defense literature makes a basic assumption that \textit{the attack source is not human}. In this paper, we challenge it by asking this question: \textit{Is it possible to attack speaker identification systems using analog manipulation of the speech signal?}

Answering this question in the affirmative has critical implications on using ML to detect and identify human speakers. An analog transform of the speech signal to evade speaker identification challenges the \textit{{identifiability}} assumption that underlies various acoustic tasks; human characteristics can no longer be uniquely identified from their speech. An attacker can control the propagation medium to affect the speaker identification task. Towards that end, we present \name, a \textit{live spoof attack}, which enables analog transformations of speech signals. \name allows the attacker to transform their voice for inducing a targeted misclassification at the \ASI system, effectively impersonating a target victim.

Realizing \name requires us to satisfy \mr{four} conditions. First, the analog transform must occur on live speech. Second, an arbitrary speaker should be able to impersonate another arbitrary victim; \ie, the attacker needs not be a professional vocalist or have any impersonation experience. Third, the transform should directly impact the \ASI model prediction. \mr{Fourth, the transform can be mathematically modeled to be incorporated in the attack optimization objective.} \name exploits the acoustic resonance phenomenon to satisfy these conditions. Acoustic resonance is a physical transform where objects vibrate to specific frequencies. Acoustic resonance allows an object to act as a frequency filter, amplifying some frequency components and dampening others.

\name uses hand-crafted tubes to apply the adversarial resonance transformation to the speaker's voice. We chose tubes as our attack's physical objects for two reasons. First, tubes are ubiquitous and inexpensive; they are available in hardware stores in different dimensions. Second, there is extensive literature on acoustic modeling of musical wind instruments, most of which have cylindrical or conical shapes. Note that the same methodology can be extended to arbitrary shapes using wave simulation and numerical analysis~\cite{allen2015aerophones, umetani2016printone}.

To realize \name, we model the tube resonator as a band-pass filter (BPF) transform; the tube dimensions fully define the filter. Next, we develop a black-box optimization procedure over the filter parameters (tube dimensions) to trick the \ASI model into recognizing the voice of a chosen target speaker. We apply an evolutionary algorithm (Sec.~\ref{sec:proposedAlg}) that uses the \ASI model to find the optimal tube dimensions for a given target. An adversary can use these parameters to realize a tube that would match their voice to a target speaker.

We perform extensive evaluation of \name on two state-of-the-art \ASI models and \mr{five} spoofing detection baselines. We validate \name on standard speaker identification dataset, VoxCeleb, and on live speech by conducting a user study of \mr{14} participants. We build a physical recording setup, and evaluate \name physically. We confirm that \name's adversarial tubes succeed in performing over-the-air impersonation attack in the real-world.

This paper makes the following contributions:
\begin{itemize}
\itemsep0em
    \item We show that a human can directly produce analog audio adversarial examples in the physical domain. This adversary bypasses current acoustic defenses based on liveness and (presumably uniquely) identifying characteristics of the speaker, such as voice pitch.

    \item We demonstrate that, using commonly available plastic tubes, an attacker can change the properties of their speech in a systematic way and manipulate ML models. For example, an adversary can impersonate 500 other speakers using tubes. Moreover, \name is only 23\% detectable by the best ASVspoof 2021 spoofing detection baseline that has 100\% accuracy on classifying natural (\ie, no tube) recordings as live.

    \item We run our attack on live speech to confirm its practicality. We perform a user study and show that the attack is successful over-the-air on live speech with 61.61\% success rate. \mr{We conduct a human impersonation study as a baseline and find that its success rate is only 6.2\%.}  %
    
    \item \mr{We discuss a set of strategies to detect the attack and add a discussion of \rev{limitations and} future work.}  
\end{itemize}

\section{Acoustics Background}

In this section, we introduce background concepts on acoustics and human speech modeling. 

\subsection{Acoustic Resonance}
\label{sec:acoustic_resonance}
Resonance is a natural phenomenon in which objects vibrate when excited with a signal that contains specific frequency components~\cite{kinsler2000fundamentals}. These frequency components are referred to as the resonance frequencies, and they contain the fundamental frequency $f_0$ (object's natural frequency) and its harmonics $f_i$. A resonating object acts as a \textit{filter} that magnifies the resonance frequencies, and filters out other frequencies in the excitation signal. The resonance vibrations encounter resistance and losses that define the filter sharpness---referred to as the quality factor $Q$. The filter's $f_0$ and $Q$ are usually well defined by the object's shape and properties. 

Acoustic resonance happens to sound waves that travel inside a hollow object, such as a tube, when it forms a standing wave~\cite{Aljalal_2015, kinsler2000fundamentals}. This phenomenon is observed in wind instruments musical notes. Similar to musical tones, human speech is produced by resonance inside the speaker's vocal structure. In \name, we exploit this phenomenon and our understanding of the human speech to design a physical speech filter using tubes and perform targeted attacks on \ASI.

\paragraph{Resonance Frequency.} 
In (cylindrical) tubes, the fundamental resonance frequency $f_0=c_{air}/\lambda$ (Hz), where $c_{air}$ is the speed of sound in air, and $\lambda$ is the standing wave wavelength. For open-ended tubes, as in our use case, the fundamental mode $\lambda=2L$ where $L$ is the tube length~\cite{nederveen1969acoustical}. Thus, $f_0 = c_{air}/2L$, and $c_{air} = 20.05 \sqrt{T}$ (m/s) in dry air~\cite{kinsler2000fundamentals}, where $T$ ($\degree$K) is the thermodynamic temperature. These equations, however, do not consider the tube diameter and air humidity. 
A more accurate equation is:
\begin{equation}
\label{eqn:f0}
    f_0 = \frac{c_{air}}{2(L+0.8d)},
\end{equation}
where $d$ is the tube diameter, and $\Delta L=0.8d$ is an empirical term derived from measurements~\cite{end_correction}. %

\paragraph{Quality Factor.} The quality factor quantifies the acoustic losses inside the tube. There are two main sources of losses~\cite{moloney2001acoustic, kinsler2000fundamentals}: radiation loss and wall loss.
The radiation loss $d_{rad}$ is the energy loss due to acoustic radiation outside the tube~\cite{kinsler2000fundamentals}: $    d_{rad} = {2\pi A f_0^2}/{c_{air}^2}$, 
where $A$ is the tube cross-sectional area. The wall losses happen because the air speed goes down to zero at the tube internal walls, hence, it leads to energy loss. Wall losses can be quantified by this damping factor~\cite{kinsler2000fundamentals}: $d_{wall} = \sqrt{\mu/\rho A f_0}$,
where $\mu=1.81*10^{-5} kg/ms$ is the air viscosity, and $\rho =1.18 kg/m^3$ is the air density. There are other losses that are either hard to quantify, or environment dependent, or can be ignored compared to the radiation and wall losses~\cite{qpipe}. Thus, the tube quality factor \rev{can be approximated by}:
\begin{equation}
\label{eqn:Q0}
    Q_0 = 1/(d_{rad}+d_{wall}).
\end{equation}

\subsection{Human Speech Modeling}
\label{sec:human_model}
\begin{figure}
  \centering
\scalebox{.9}{
     \begin{subfigure}[b]{0.4\columnwidth}
         \centering
         \includegraphics[width=\textwidth]{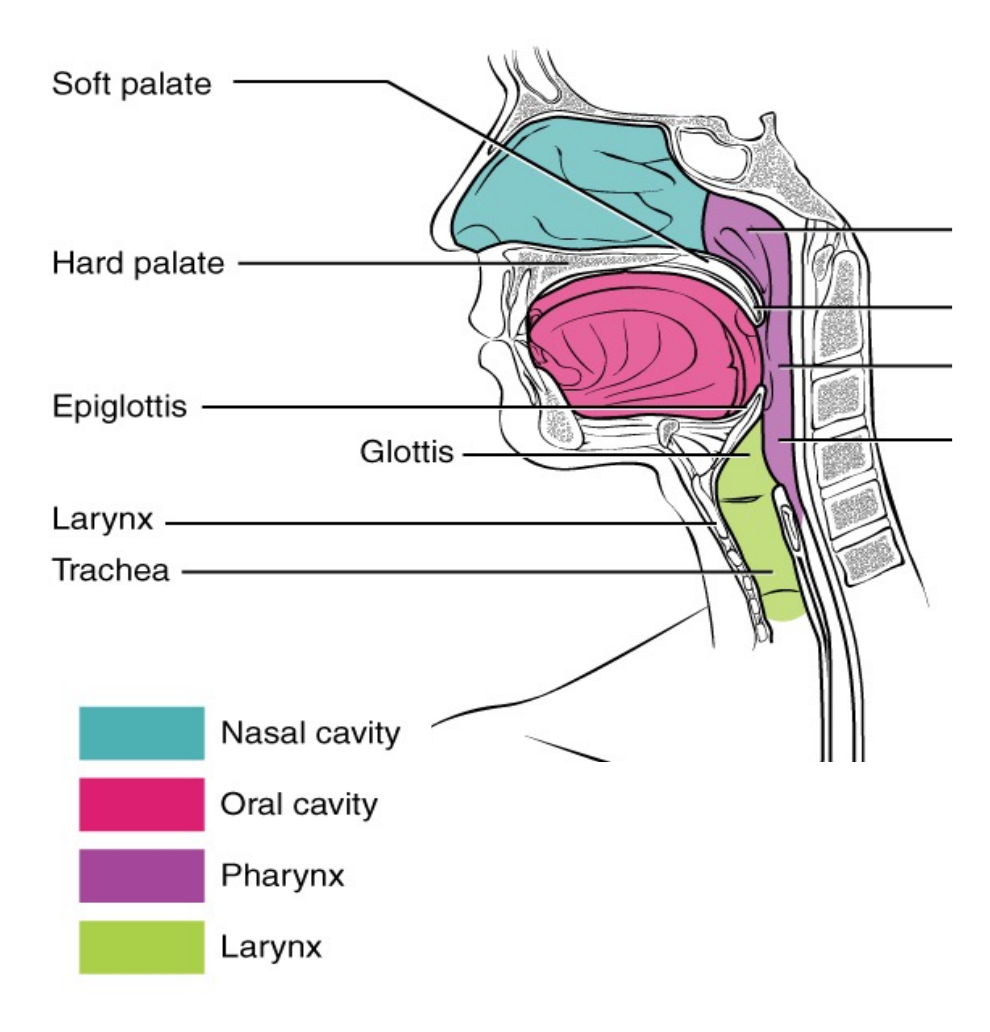}
         \caption{Vocal Tract Structure.}
         \label{fig:vocal_tract}
     \end{subfigure}
     \begin{subfigure}[b]{0.55\columnwidth}
         \centering
         \includegraphics[width=\textwidth]{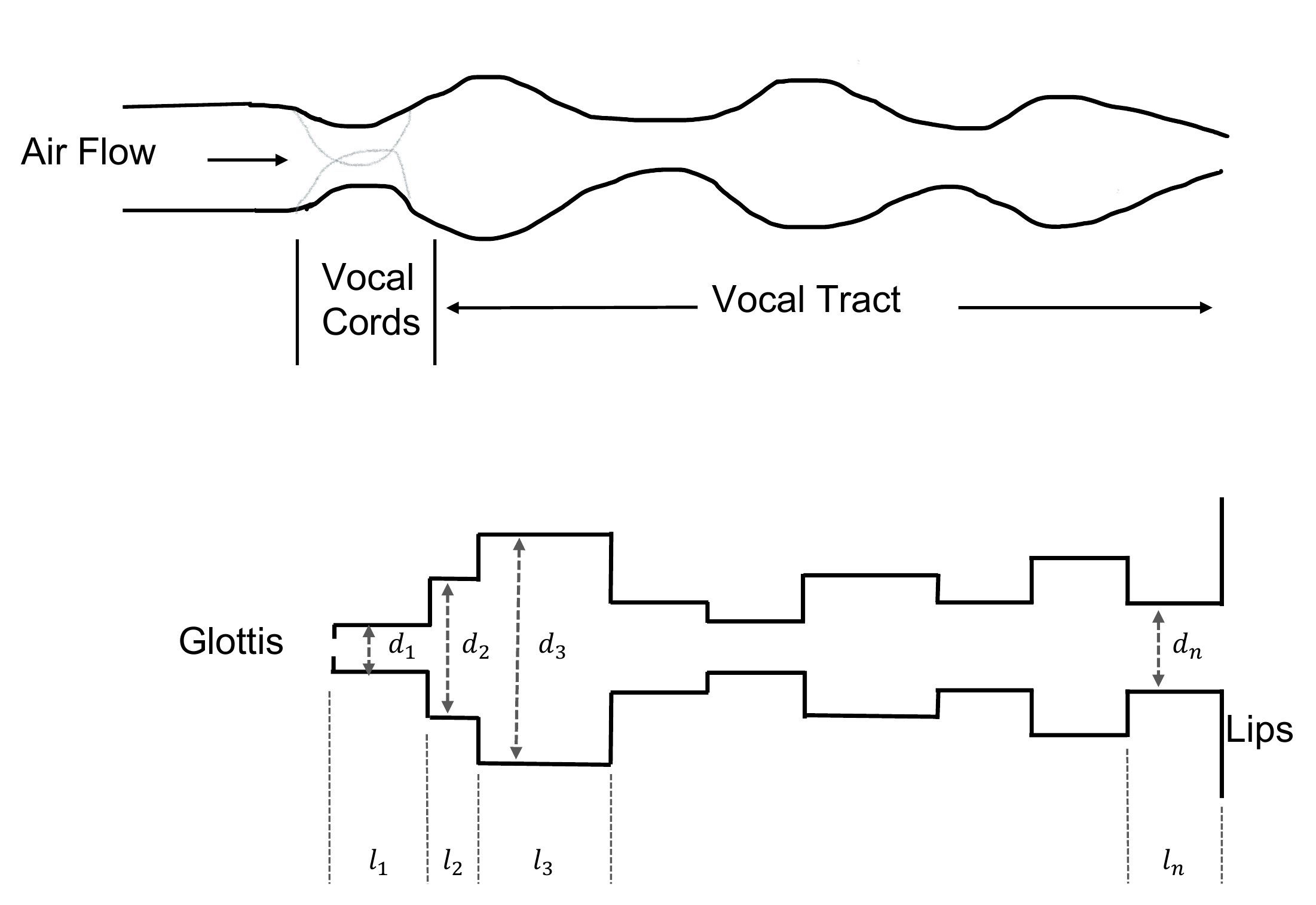}
         \caption{Vocal Tract Model.}
         \label{fig:vocal_tubes}
     \end{subfigure}}
     \caption{The vocal tract structure and model. (a) The structure including the glottis, the pharynx, the oral cavity, the nasal cavity, and the lips---adapted from AnatomyTool~\cite{anatomytool}.  (b) Vocal tract parts modeled as consecutive tubes of different diameters.}
     \label{fig:vocal_tract_and_tube_model}
\end{figure}

\paragraph{Biological Characteristics.}
Humans generate speech using three main structures~\cite{stevens1998acoustic}: the lungs, the vocal folds (glottis), and the articulators as shown in Fig.~\ref{fig:vocal_tract}. The lungs produce airflow and control air pressure, this airflow in turn makes the vocal folds vibrate and modulate the passing air to produce sound (audible air vibrations)---referred to as the glottal excitation. The vocal folds physical shape controls the vibrations frequency, hence, it is considered the \textit{speech source}~\cite{stevens1998acoustic}.
The vibrating air passes through the articulators---referred to as the vocal tract---such as the pharynx, the oral cavity, the tongue, the nasal cavity, and the lips. The vocal tract forms a flexible airway that shapes the sound into the final distinctive speaker voice. The moving parts, such as the tongue and lips, change their position to produce different sounds and speech phonemes. Thus, the vocal tract is considered a \textit{linear acoustic filter}~\cite{stevens1998acoustic}, and human speech production is  modeled as a sound source followed by an acoustic filter.

\paragraph{Source-Filter Model.}
The glottal excitation defines the voice \textit{pitch} and can be modeled by an impulse train in the time domain $g(t)$ and by harmonics in the frequency domain $G(f)=\mathcal{F}(g(t))$. The vocal tract can be modeled as a variable acoustic resonator $H_v(f)$ that filters the glottal excitation into speech $s(t) = \mathcal{F}^{-1}(H_v(f) \cdot G(f))$. The resonator characteristics depends on the vocal tract size and shape; \ie~the speaker's anatomy, and the speech phonemes vary with the tongue and lips movement. The different parts of the vocal tract are modeled as consecutive tubes~\cite{5311870}, as shown in Fig.~\ref{fig:vocal_tubes}. The tubes are an acoustic resonator that amplifies certain frequencies and filters out others to shape the acoustic excitation into a specific voice and speech sound.

\section{System and Threat Models}
\label{sec:sys_threat_model}

In this paper, we consider Automatic Speaker Identification (\ASI)---a classification task that determines a speaker's identity, based on their speech~\cite{snyder2017deep}, from a set of enrolled speakers. Typically, the identification task can be text-dependent; \ie~the speaker has to say a predefined utterance, or text-independent; \ie~the speaker can say any utterance of their choice. Text-independent \ASI \mr{ is more secure against replay attacks, and more usable as it can be embedded within other tasks such as speech recognition in a seamless interaction.} %

\paragraph{System Model.} We consider a system that applies the \ASI task for user identification and authentication. The system collects speech samples from its users during the enrollment phase to extract their voiceprint (speaker embeddings) and fine-tune the \ASI model. 

Modern \ASI systems are based on speaker embedding by deep neural networks. These models  capture the \mr{speaker's voice} characteristics from a variable-length speech utterance $s(t)$ and map it to a vector (embedding) in a fixed-dimensional space. X-vector DNN~\cite{snyder2017deep, snyder2018} is a common \ASI embedding network which consists of 3 stages: (1) feature extraction, (2) speaker embedding, and (3) classification. The first stage extracts the mel-frequency cepstrum coefficients (MFCC) which reduce the dimensionality of the speech signal into a 2D temporal-spectral map, and applies voice activity detection (VAD) to filter out non-speech segments. Second, a time-delayed neural network (TDNN) maps the variable-length MFCC samples into a fixed-dimensional embedding (x-vectors) space. Finally, a softmax layer is applied on x-vectors to obtain the predicted identity of the speaker. The network is trained using a multi-class cross entropy objective.

During inference, the system \mr{collects a speech utterance from the user}, and runs the \ASI task to determine the user's identity. The \ASI task is the \textit{only} access control mechanism deployed by the system. The system also applies a spoofing detection technique as a countermeasure against spoofing attacks; as we detail next in the threat model as well as Sec.~\ref{sec:related}. 

Fig.~\ref{fig:overview} shows the system setup. The system runs a spoofing detector that determines whether the recorded utterance is from a live speaker or digitally produced, i.e., spoofed. If the utterance is detected to be live, the spoofing detector feeds it to the \ASI model which classifies the speaker identity and grants the user access to the secure system.
This system setup can be deployed for logical access applications such as phone banking services, voice assistants, and smart home devices.

\paragraph{Threat Model.} We consider an adversary that wants to attack the \ASI model to be identified as a target user. First, the adversary will not perform conventional spoofing techniques such as replay, speech synthesis, voice conversion, or digital adversarial examples to evade detection by the system's spoofing detector. Note that spoofing detection techniques (Sec.~\ref{sec:related}) are based on the assumption that spoofed speech is always generated by a \textit{digital} speaker, not a live human. Instead, the adversary will \textit{naturally} impersonate the victim's voice by changing their \textit{live} voice using physical objects. Our work introduces a systematic reproducible technique that allows the adversary to impersonate an arbitrary speaker's voice\mr{, in the eyes of the \ASI model,} without using a digital speaker. The attack is analog and only allows for the use of physical objects and natural sounds. 

 \mr{Second, the adversary performs an audio-only interaction with the system.
 Hence, they have complete control over the recording environment, as shown in Fig.~\ref{fig:overview}.  They have no access to the \ASI model internals; \ie, a black-box attack. The adversary can only query the \ASI model on inputs of their choice and get the model's output scores and label. As such, the adversary needs no recordings of the victim's speech. They only know the victim is enrolled in the \ASI model. Finally, the adversary impersonates the victim in the eyes of the \ASI model to gain access to their protected accounts. The attack does not target human listeners explicitly. }

\section{Attack Methodology}
\label{sec:method}

This section introduces our attack, \name, provides a theoretical intuition, and details its operation. 

\subsection{Overview}

Fig.~\ref{fig:overview} displays \name's system and attack flow. A microphone captures the speaker's voice\mr{, validates the voice liveness,} and feeds it to an \ASI system. \name exploits the flawed assumption that spoof attacks must be generated from a digital speaker. The current \ASI setup overlooks the acoustic environment attack vector. \name challenges these assumptions and performs an attack that is live by default. An attacker speaks through a specifically designed tube to induce a targeted misclassification at the \ASI system, effectively impersonating a target victim.

\paragraph{Attack Description.} The attack is as follows. The adversary models the tube resonator as a band-pass filter (BPF) transform (Sec.~\ref{sec:explain_resonance}). The filter is fully defined by the tube dimensions. Next, the adversary runs an optimization function over the filter parameters (tube dimensions) to trick the \ASI model into classifying the voice as a chosen target speaker. In a black-box setting, we apply an evolutionary algorithm (Sec.~\ref{sec:proposedAlg}) that uses the \ASI model score and label to find the optimal tube dimensions for a given target speaker:
\begin{align}
\label{eq:advobj}
\small
\centering
    \min_p &\quad \operatorname{R}(\operatorname{\ASI}(s'), y_t) \quad \text{s.t.} \quad s' = \operatorname{F_{tube}}(s, p),
\end{align}
where $s$ is the original speech sample, $p$ is the tube parametrization, $y_t$ is the attack target label, $R$ is the loss, $\operatorname{F_{tube}}(.)$ is the mathematical model of the tube, and $\operatorname{\ASI}(.)$ is the model under attack. The adversary would then purchase the required tube, and speak through it to trick the system. Therefore, the adversary is able to systematically bypass spoofing detection and attack \ASI with an analog attack.

\subsection{Modeling Resonance in Tubes}
\label{sec:explain_resonance}

Modeling the filter corresponding to a particular tube is a key requirement for \name. We model the tube transfer function $H_{res}(f)$ as a sum of band-pass filters (BPFs), with a filter at each harmonic. The $i^{th}$ filter $H_i(f)$ is defined by its center frequency at the resonance harmonic $f_i$, and the filter width $\Delta f_i$ is defined by the quality factor $Q_i$ (\eqref{eqn:fQi}), where $i=1,2, \cdots, \lfloor f_s/f_0\rfloor$ is the harmonic number, and $f_s$ is the speech sampling rate. The input speech signal $s_{in}(t)$ resonates at the tube's fundamental frequency $f_0$ and its harmonics $f_i=i \cdot f_0$.  Thus, the tube output speech signal is:
\begin{equation}
    \label{eqn:outputSignal}
    s_{out}(t) = \operatorname{F_{tube}}(s_{in},p) = \mathcal{F}^{-1} (H_{res}(f) \cdot S_{in}(f)),
\end{equation} 
where $\mathcal{F}^{-1}$ is the inverse Fourier transform, $S_{in}(f) = \mathcal{F}(s_{in}(t))$ is the input speech spectrum, $H_{res}(f)=\sum{H_i(f)}$ is the tube transfer function, and $p=(L,d)$ are the tube parameters. Note that $H_{res}(f)$ is parameterized by $p$, but we drop this parameterization to make the notation simpler. In \name, we adopt a simple two-pole band filter for $H_i(f)$.

\paragraph{Single Tube.} 
Given a single tube with length and diameter parameters $p$, Eqn.~\ref{eqn:f0} and~\ref{eqn:Q0} quantify the fundamental resonance parameters. The full harmonic range of $f_i$ and $Q_i$ are:
\begin{equation}
\label{eqn:fQi}
\begin{gathered}
    f_i = i\cdot f_0 = \frac{i\cdot c_{air}}{2(L+0.8d)};\quad 
    Q_i = Q_0/\sqrt[4]{i}, 
\end{gathered}
\end{equation}
where $i$ is a positive integer representing the harmonic number for open-ended tubes.

Our lab measurements revealed that there is about 1\% mismatch between the theoretical (Eqn.~\ref{eqn:f0}) and measured $f_0$.  We attribute this mismatch to the end-correction term uncertainties and air humidity. Also, we estimated $Q_i$ empirically, as its change with $f_i$ depends on the dominating loss for a given tube. We found that $Q_i$ decays as $1/i$, $1/\sqrt{i}$, or $1/\sqrt[4]{i}$ give reasonable estimates and we decided to select the latter. We include both corrections in the filter formulation.

\paragraph{Multiple Tubes.}
Next, we extend the single tube model into a structure of multiple consecutive tubes of different lengths and radii to increase \name's degrees of freedom and the set of possible filters. The extended structure can reach a wider range of spoofed identities, hence, it increases the attack success rate as shown in Sec.~\ref{sec:91_attack_scaled}.  

Resonance inside connected open-ended tubes happens when the acoustic impedance between the connected tubes equal an open-end impedance~\cite{stevens2000acoustic}. This condition is mapped to the following equation for each two tubes intersection:
\begin{equation}
\label{eqn:2tubes_nonlinearEq}
    A_1 \cdot \cot(2\pi f L_1/c_{air}) = A_2 \cdot \cot(2\pi f L_2/c_{air}), 
\end{equation}
where $A_1$ and $A_2$ are the two tubes cross-sectional areas, $L_1$ and $L_2$ are their lengths. We solve this non-linear equation numerically to obtain the resonance frequencies $f_i$'s.

\begin{figure*}[t]
  \centering
  \scalebox{0.9}{
     \begin{subfigure}[b]{0.24\textwidth}
         \centering
         \includegraphics[width=\textwidth]{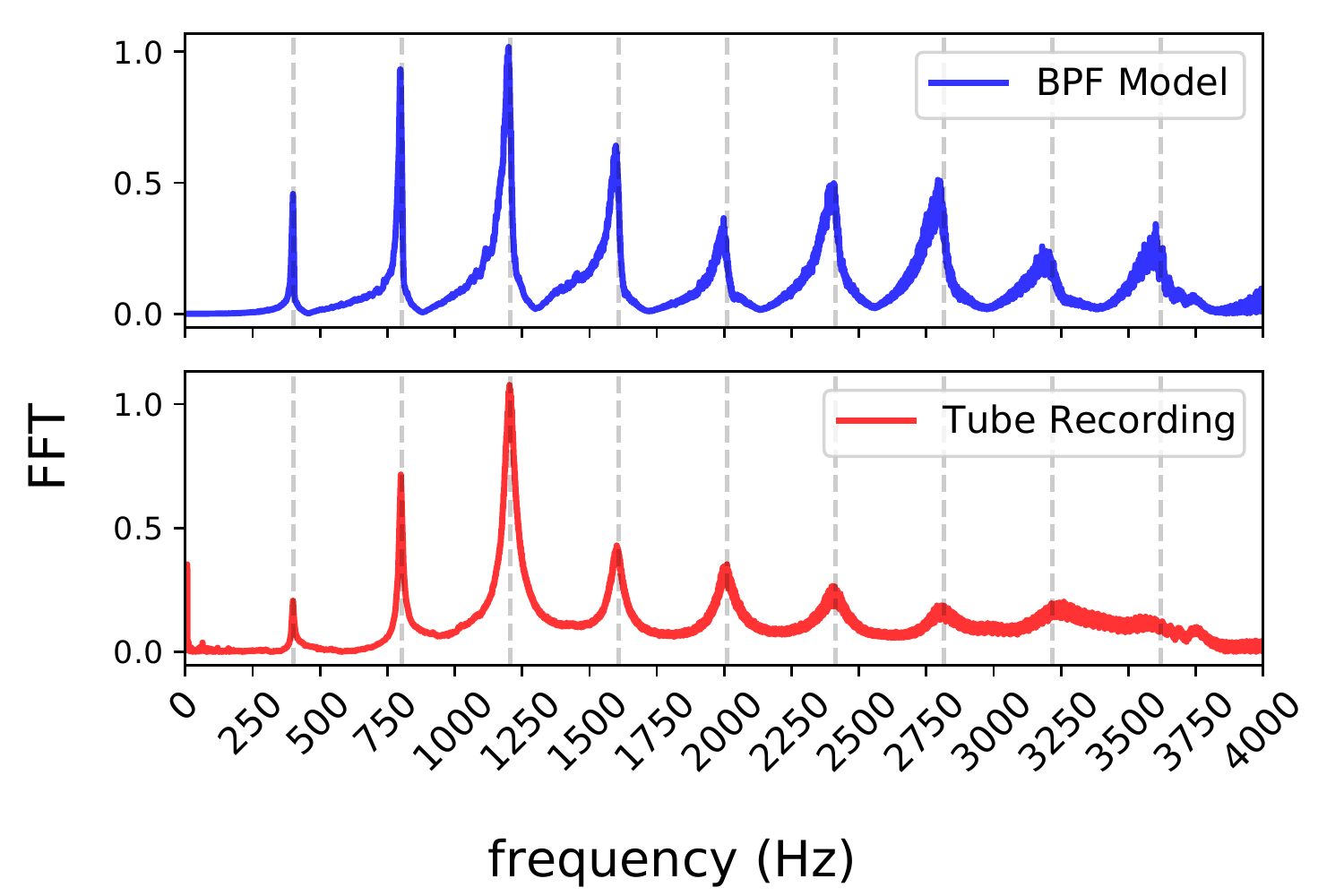}
         \caption{FFT of a chirp}
         \label{fig:chirp_fft_16}
     \end{subfigure}
     \begin{subfigure}[b]{0.24\textwidth}
         \centering
         \includegraphics[width=\textwidth]{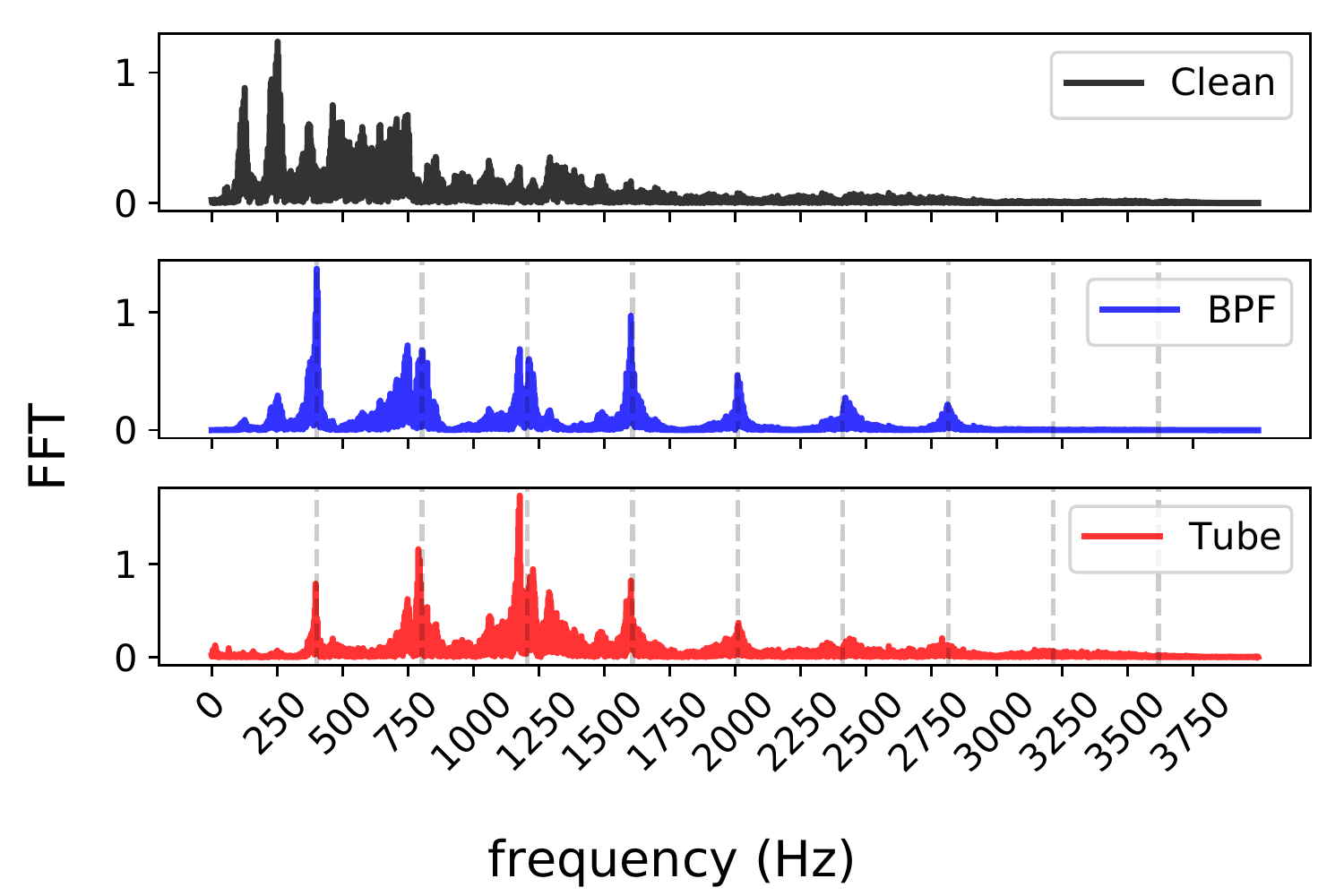}
         \caption{FFT of a speech sample}
         \label{fig:speech_fft_16}
     \end{subfigure}
     \hfill
     \begin{subfigure}[b]{0.24\textwidth}
         \centering
         \includegraphics[width=\textwidth]{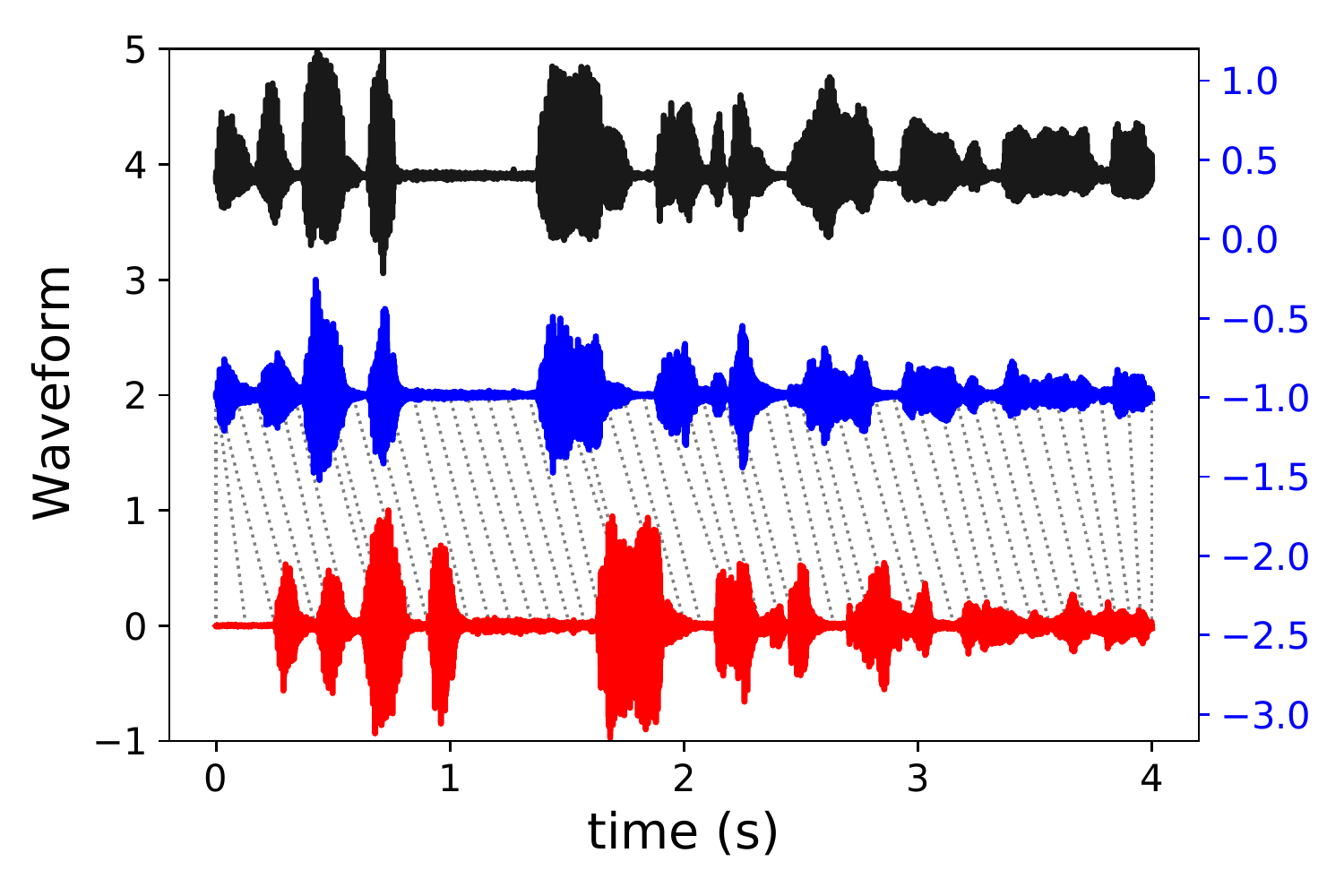}
         \caption{Speech waveform}
         \label{speech_dtw_16}
     \end{subfigure}
     \hfill
    \begin{subfigure}[b]{0.23\textwidth}
     \centering
     \includegraphics[width=\textwidth,height=26mm, keepaspectratio]{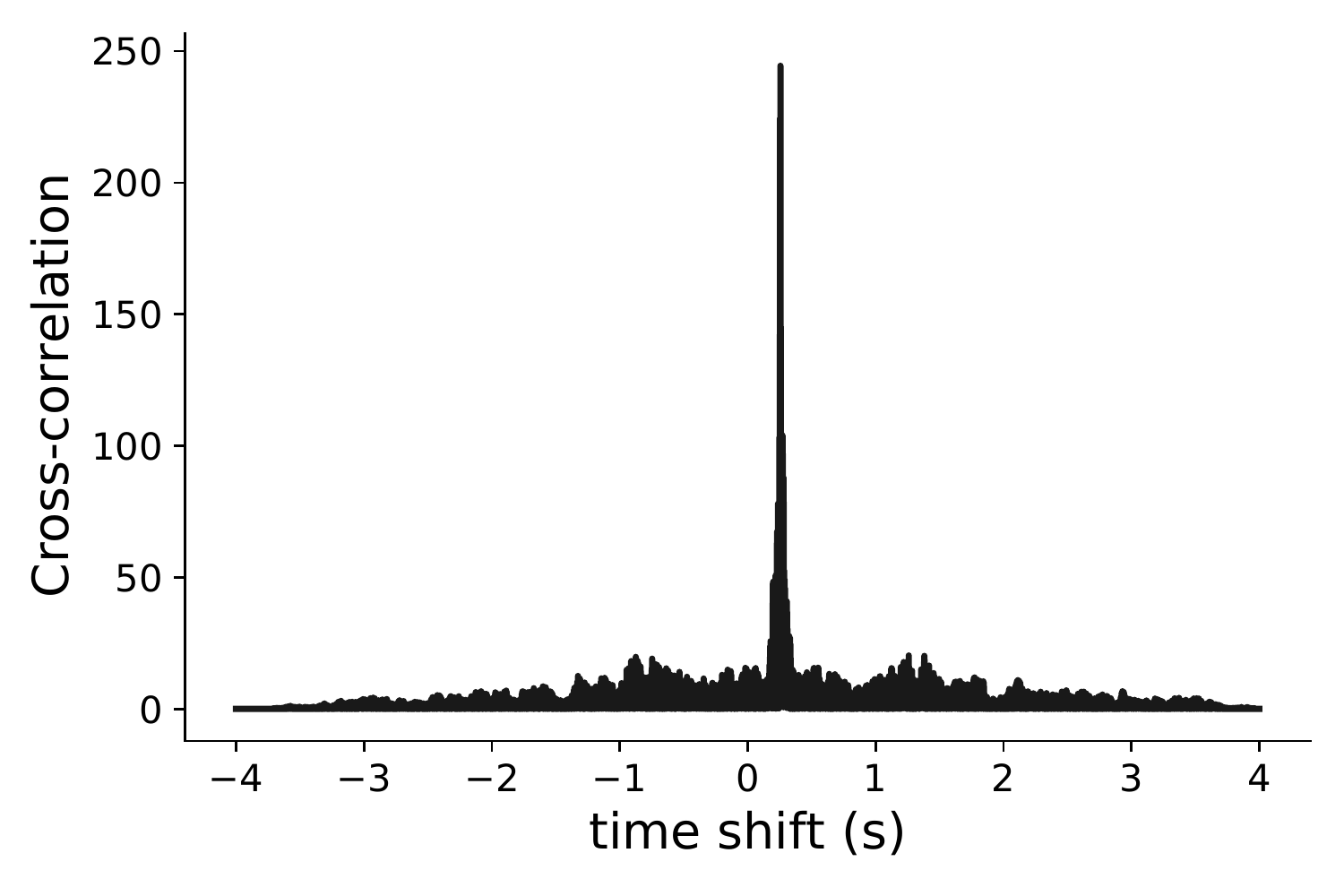}
     \caption{Cross-correlation between the tube and BPF waveforms}
     \label{cross_corr_16}
     \end{subfigure}}
     \caption{Resonance model validation of Tube 1 ($L=40.6, d=3.45$) vs its BPF model: (a) FFT of chirp, (b) FFT of a speech utterance, (c) speech wavefroms showing DTW alignment between tube and BPF signals, (d) cross-correlation between tube and BPF waveforms.} 
     \label{fig:resonance_speech}
\end{figure*}

\paragraph{Validation.}
We validate the resonance model by measuring real tubes resonance and comparing it to our BPFs model. \mr{First, we excite the tube with a 3-second chirp signal~\cite{smyth2009estimating} that exponentially spans the frequency range from 100 to 3700 Hz. Then, we play speech samples from VoxCeleb dataset and measure the similarity between tube and BPF output signals. We \rev{use} the setup in Fig.~\ref{fig:box_setup} for recording.}

\mr{Fig.~\ref{fig:resonance_speech} shows the Fast Fourier Transform (FFT), waveform, and cross-correlation plots for a tube of $L=40.6,\: d=3.45$ cm\ifpaper
\else
.
\fi
Fig.~\ref{fig:chirp_fft_16} shows the FFT of the chirp output, which is effectively the tube's transfer function $H_{res}(f)$. The vertical dotted lines indicate the theoretical resonance frequencies, $f_i$, which align perfectly with the measurement. Fig.~\ref{speech_dtw_16} shows the waveforms with the dynamic time warping (DTW) alignment, and Fig.~\ref{cross_corr_16} shows that the waveforms are highly correlated. We also measure the DTW alignment distance for a set of 6 tubes (Table~\ref{tab:box_matching}), which is a measure of similarity. The distances are 0.027, 0.03, 0.025, 0.023, and 0.021. Thus, the tube and BPF waveforms are very similar for all evaluated tubes. Therefore, the BPF model is a realistic representation of the tube resonance. The attacker uses this model to obtain the tube parameters for a targeted attack.}

\subsection{Attack Intuition}
\label{sec:intuition}
Speech technology applications such as speech recognition, speaker identification, and keyword spotting are highly sensitive to the acoustic environment. Models trained on clean speech recordings often fail in real world scenarios~\cite{ko2017study, pervaiz2020incorporating, hu2018generative}. Usually, training data has to be augmented with simulated environmental effects such as noise and echo~\cite{ko2017study, pervaiz2020incorporating, hu2018generative}. The same applies for speech adversarial examples. Adversarial perturbations do not succeed over-the-air when the environmental variations are not considered in the optimization objective~\cite{qin2019imperceptible, ahmed2022towards}. Hence, one of the fundamental intuitions behind \name is that if the acoustic environment falls outside the expected distribution, the model predictions will become unreliable. 

Still, one can wonder why a tube (resonator) has such a high impact on the \ASI model's performance. In \Cref{sec:pitch_shift}, we theoretically show that tubes affect the estimated pitch. Next, we empirically validate that tube parameters are statistically significant predictors of pitch shifts between input and output signals. 
Such pitch shifts introduce distribution shifts w.r.t the real-world utterance datasets used to train speech models. It has been well-established that such distribution shifts reduce model performance at inference time~\cite{quinonero2008dataset,sugiyama2012machine}. In particular, \ASI is sensitive to the pitch of the speech signal; therefore, applying the tube is expected to alter the classification. 

\subsubsection{Tubes Cause Pitch Shifts}
We build on the work of McAulay and Quatieri~\cite{pitch_estimation} who frame the pitch estimation as the solution of an unconstrained optimization of the mean square error between the Short-time Fourier transform (STFT) of a signal $s(t)$ and a sum of harmonics, parameterized by the pitch. In  \Cref{sec:pitch_shift}, we show that the resonance effect of the tube translates to a constrained version of the same optimization problem. We then argue that given the smaller feasibility set of the constrained problem, its solution will inherently have filtered out frequencies. As a result, the estimated pitch will be different.

\paragraph{Validation.} We design an experiment to study the correlation between the pitch shift and the change in the classification result. We played samples from the VoxCeleb dataset through three tubes of different lengths (corresponding to different resonance frequencies). For each sample, we estimated the pitch of both signals (original and output) using CREPE~\cite{crepe} which provides a time-domain signal of the signal pitch. Given that the pitch varies in the duration of each utterance, we need to account for different speakers, utterances and original clip recordings to establish a generalized relationship between pitch shifts and tube parameters.

We regress this pitch difference using an ordinary least squares model with a design matrix containing tube parameters and 2060 audio samples. The linear regression model achieves an $R^2=0.552$. Therefore, the tube parameters explain at least 55\% of the pitch shift variances. P-values achieved are $1.77\times10^{-26}$ and $2.99\times 10^{-149}$ for length and parameter, respectively, which means that these tube parameters are good regressors of the shifts introduced by the tube in a variety of recording conditions, utterances and speakers.

\subsection{\name's Algorithm}
\label{sec:proposedAlg}

\begin{algorithm}[t]
\small
\caption{Differential Evolution}
\label{alg:de_algo}
\begin{algorithmic}[1]
        \STATE {\bfseries Input:} $s$, $y_t$, pool size $N$, attack budget $n$, fitness function $f$, crossover parameter $c$, maximum iterations $it$, mutation proportion $m$

        \STATE $A : N\times n = \text{\textsf{random}(pool)}$
        \FOR{$i=0$ {\bfseries to} $it$}
            \STATE $A_{new} : N\times n = 0.0$
            \FOR{$j=0$ {\bfseries in} $N$}
                \STATE $r_1, r_2 = \text{\textsf{sample-randomly}}(A)$
                \STATE ${l} = A_{best} + m \times (r_1 - r_2)$
                
                \STATE $m = c > \text{\textsf{random-mask-of-size}}(n)$
                \STATE $a = l * m + A_j * (1-m)$
                
                \IF{$f(a,s,y_t) > f(A_j,s,y_t)$}
                    \STATE $A_{new,j} = a$
                \ELSE
                    \STATE $A_{new,j} = A_j$
                \ENDIF
            \ENDFOR
            \STATE $A = A_{new}$
        \ENDFOR
\end{algorithmic}

\end{algorithm}

\begin{figure}[t]
    \centering
    \scalebox{0.7}{
    \includegraphics[width=\columnwidth]{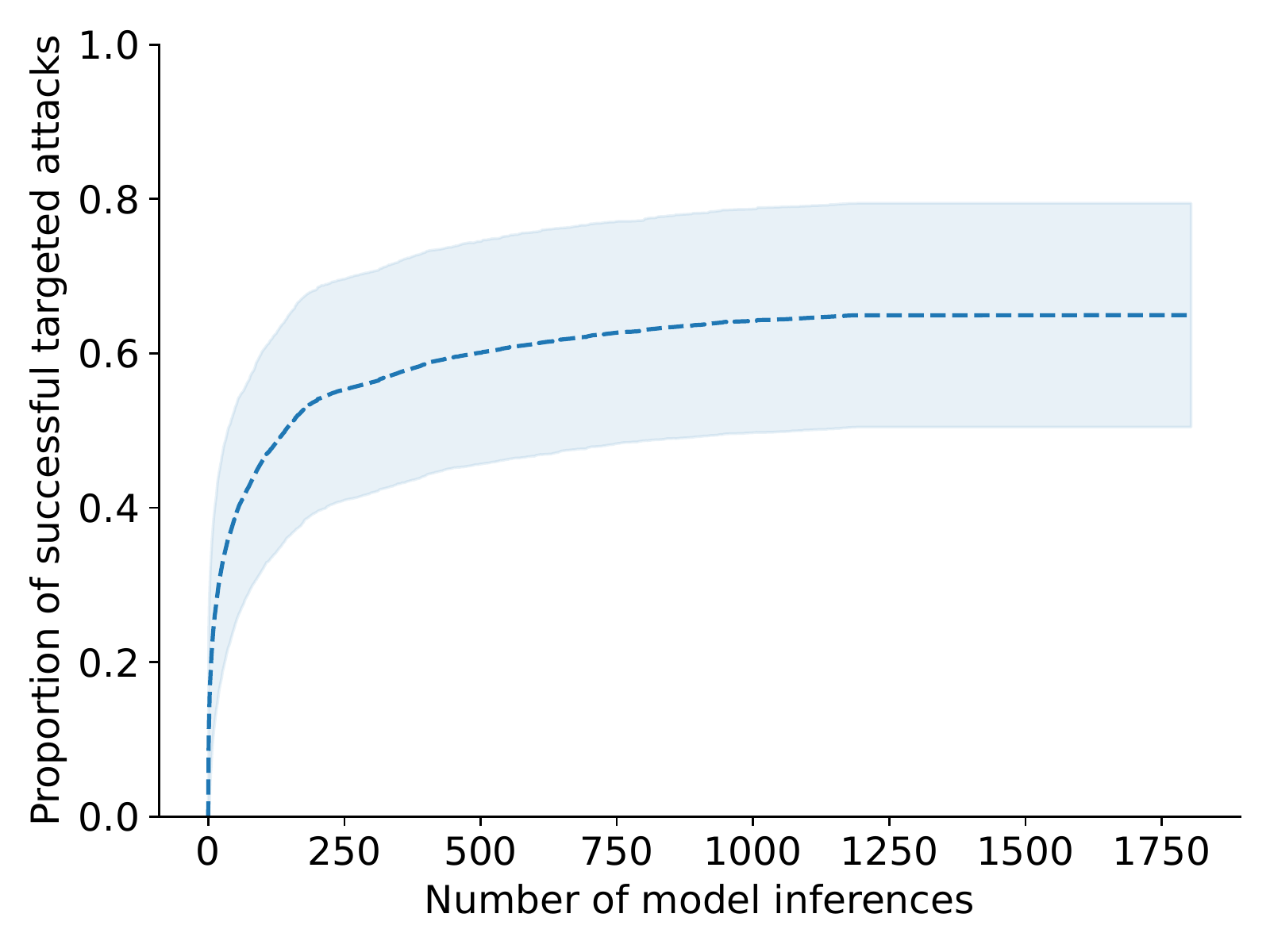}}
    \caption{Average reachable target search performance across all of the participants with SpeechBrain model}
    \label{fig:meta_results}
    \vspace{-0.3cm}
\end{figure}

In Sec.~\ref{sec:explain_resonance}, we parameterize the tubes by the quality factor $Q_0$ and the fundamental frequency $f_0$. Although, for a single-tube configuration, the search space is small enough to be bruteforced within a few minutes, we find that in many cases we can speed up the attack using optimization. More precisely, we experiment with gradient-free non-convex optimization algorithm from a family of evolutionary algorithms called \textit{differential evolution} (DE)~\cite{storn1997differentialevolution}.
\mr{Algorithm~\ref{alg:de_algo} describes our DE approach with \textit{best2exp} strategy. The algorithm performs the tube parameters $A$ search by picking three data samples from an underlying population and combining the best performing one with the difference between the other two.
The algorithm is called differential, since the update step includes computing the difference between a pair of samples and stochastically appending it to the third. In the search algorithm, we set boundary conditions on the tube dimensions which defines the underlying population. We define the boundaries as: $f_0$ ranges from 50 Hz to 1 kHz, and its $Q_0$ ranges from 5 to 100, such that $f_0$ falls in the typical range of human voice pitch. We sample from this range using step size of 10 Hz for $f_0$ and 5 for $Q_0$. According to~\eqref{eqn:f0} and ~\eqref{eqn:Q0}, the single tube length would range from 10 cm to 3 m, and the diameter ranges from 1 cm to 15 cm, which is a practical range. For two-tube structures, each tube length can range from 5 cm to 120 cm with 5 cm step size, and the areas ratio ranges from 1 to 10 with step size of 1. The resultant $f_i$'s are found from \eqref{eqn:2tubes_nonlinearEq}. We set the population size $N=100$, maximum iterations $it=5$, and tolerance of 0.001. The attack is performed in a black-box fashion, requiring only the target class score of the \ASI model. Thus, the fitness function $f(A,s,y_t)$ is the \ASI model's score of the target label $y_t$ when transformation $A$ is applied on the user's utterance $s$.} We find that within 100 model invocations, as is demonstrated in Fig.~\ref{fig:meta_results}, we could find $46\%\pm12$ of all possible reachable targets, whereas at 250 invocations it grows to $55\%\pm14$. Despite relatively low performance, our DE algorithm enables the attacker to within minutes check with a reasonable probability if a user's utterance $s$ can be transformed to impersonate a target $y_t$. 

\section{Experimental Setup}
\label{sec:exp_setup}
 
We design an experimental setup, comprising speech datasets, ASI models, spoofing detection models, and a physical measurement setup to evaluate our proposed attack, \name. Our evaluation answers the following questions:

\begin{itemize}
\itemsep0em
\item[\textit{Q1.}]\textit{ How well does \name perform as an impersonation attack on \ASI models? }
We instantiate \name on a standard dataset, VoxCeleb, using the resonance filter model. We show that \name can successfully attack two \ASI models. Using \name, each adversarial speaker successfully impersonates 500 targeted victims, on average. (Sec.~\ref{sec:91_attack_scaled})

\item[\textit{Q2.}] \textit{Does \name's impersonation succeed in real-world?}
We build a physical recording setup and run \name over-the-air on VoxCeleb (Sec.~\ref{sec:over-the-air}). We also conduct a user study and evaluate \name on live speech. \mr{We show that \name's attack success rate over-the-air is 61\% on a standard dataset and 61.61\% on live speech. We also compare \name against human impersonation as a baseline and find that most participants were not able to reliably impersonate a target speaker with a success rate of 6.2\% on average. Finally, we show that \name is consistent over multiple trials.}

\item[\textit{Q3.}] \textit{\mr{How can a defender detect \name?}}
\mr{We study different strategies to detect \name. We show that
while the \name-generated and victim voiceprints are similar, 
the \ASI model is less confident under \name. Further, we show that a human can discern samples generated from \name. Finally, we find that \name is successful against baseline spoofing detection, but not against \rev{a detector trained on \name's samples.}}
\end{itemize}

\subsection{Datasets and ML Models}
\label{sec:models_implemention}

\paragraph{\ASI Models.} We evaluate two state-of-the-art ASI models: (1) the x-vector network~\cite{snyder2018} implemented by Shamsabadi et al.~\cite{shamsabadi2021foolhd}, and (2) the emphasized channel attention, propagation and aggregation time delay neural network (ECAPA-TDNN)~\cite{desplanques2020ecapa}, implemented by SpeechBrain.\footnote{SpeechBrain (\url{https://github.com/speechbrain/speechbrain/}) is an open-source state-of-the-art toolkit on \href{https://huggingface.co}{Hugging Face}} Both models were trained on VoxCeleb dataset~\cite{Nagrani17, Nagrani19, Chung18b}, a benchmark dataset for \ASI. The x-vector network is trained on 250 speakers using 8 kHz sampling rate. ECAPA-TDNN is trained on 7205 speakers using 16 kHz sampling rate. Both models report a test accuracy within 98-99\%.

\paragraph{Evaluation Dataset.} Both \ASI models are trained on VoxCeleb. Thus, we use VoxCeleb as our test dataset. We select a subset of 91 speakers, 45 female and 46 male speakers, that are common in the training dataset of both models. We select 20 random utterances per speaker on which both models achieve 100\% accuracy.

\paragraph{Spoofing Detection Models.}
\mr{We evaluate two spoofing detection techniques, (1) ASVspoof baselines and (2) Void. We consider two state-of-the-art baselines from the ASVspoof 2021 challenge\footnote{\url{https://www.asvspoof.org}} for physical access (PA) and logical access (LA) tasks. The PA task objective is to discriminate between live-human speech and replayed recordings via loud speakers, while the LA task objective is to differentiate between live speech and artificially generated speech using text-to-speech, voice conversion, or hybrid algorithms. The LA task considers only logical attacks; \ie~the adversary feeds the spoofed utterance digitally to the \ASI model and does not play it over-the-air. Thus, the PA and LA tasks are designed to distinguish two different features of spoofed speech: loud speakers artifacts, and synthetic speech artifacts. We use the official implementation\footnote{\url{https://github.com/asvspoof-challenge/2021}} employing the light CNN (LCNN) model~\cite{wang2021comparative}. However, each is trained on a task specific dataset from ASVspoof 2019 challenge: \textit{bonafide} and \textit{replayed} samples for the \rev{PA-LCNN} model, and \textit{bonafide} and \textit{synthetic} samples for the \rev{LA-LCNN} model. %
The second spoofing detection technique is Void (\underline{Vo}ice L\underline{i}veness \underline{D}etection)~\cite{ahmed2020void}, a recent high-performing system that uses spectral analysis to detect synthetic speech. It extracts 97 spectral features to train an SVM model. The key assumption is that live speech power is higher at low frequencies than at high frequencies, while the synthetic speech power is linearly spread out across the frequency range. This makes Void a good candidate for detecting \name since the resonance effect redistributes the speech power and amplifies the power at $f_{0}$ and its harmonics $f_{i}$ as shown in Fig.~\ref{fig:resonance_speech} and Sec.~\ref{sec:pitch_shift}. We use Wenger et al.'s implementation~\cite{wenger2021hello}, where they train three models on the ASVspoof dataset: (1) SVM, (2) Light CNN~\cite{lavrentyeva2017audio}, and (3) a custom 5-layer CNN.}

\paragraph{{Live Human Impersonation.}} We conduct a user study to test \name on live speech, involving \mr{three} stages.\smallskip

\noindent
-- In the first stage, each participant records the \rev{first 50 utterances of the arctic dataset}\footnote{\url{http://www.festvox.org/cmu_arctic/}} using a microphone, without a tube.  \rev{Since \ASI is a text-independent task, we did not place any requirements or assumptions on the utterances’ linguistic content. The use of the arctic dataset is an arbitrary choice.}
We then \mr{apply \name on these recordings to impersonate victims enrolled in the \ASI models---speakers from VoxCeleb.} %

\noindent
-- In the second stage, \mr{we validate \name's success rate by conducting the attack over-the-air. We select three representative tubes that are common between the impersonation attacks of all participants.} We ask each participant to speak each utterance through each tube and compare the live classification result to the one obtained from the filter. \mr{We ask the participants to maintain the same speaking style and not to press their lips against the tube opening as it creates non-linear transformations not captured by \name's model.} %

\noindent
-- \mr{In the third stage, we ask the participants to impersonate from 1 to 8 target speakers, based on their capacity. We select the targets from the successful impersonations using \name. Each participant watches videos of the target (celebrity) speaker till they feel confident about impersonating them, which took from 5 to 20 minutes each. Then, the participant is allowed five attempts to impersonate the target using their own words; i.e. they were not given a specific script to read.} 

We recruited 14 individuals\footnote{Two participants abstained from conducting the third stage of the study.} (7 males, 7 females, age:18-30). We obtained IRB approval from our institution to conduct the study. We collected no personal information, obtained informed consent from each participant, and followed health protocols. We use the ASI models described above, without retraining as to mimic a realistic attacker, which would attack black-box models. We use the physical setup, described below, to conduct the user study.

\subsection{Physical Setup for the Attack}
\label{sec:physical_setup}
\begin{figure}[t]
    \centering
    \scalebox{0.8}{
  \includegraphics[width=\columnwidth]{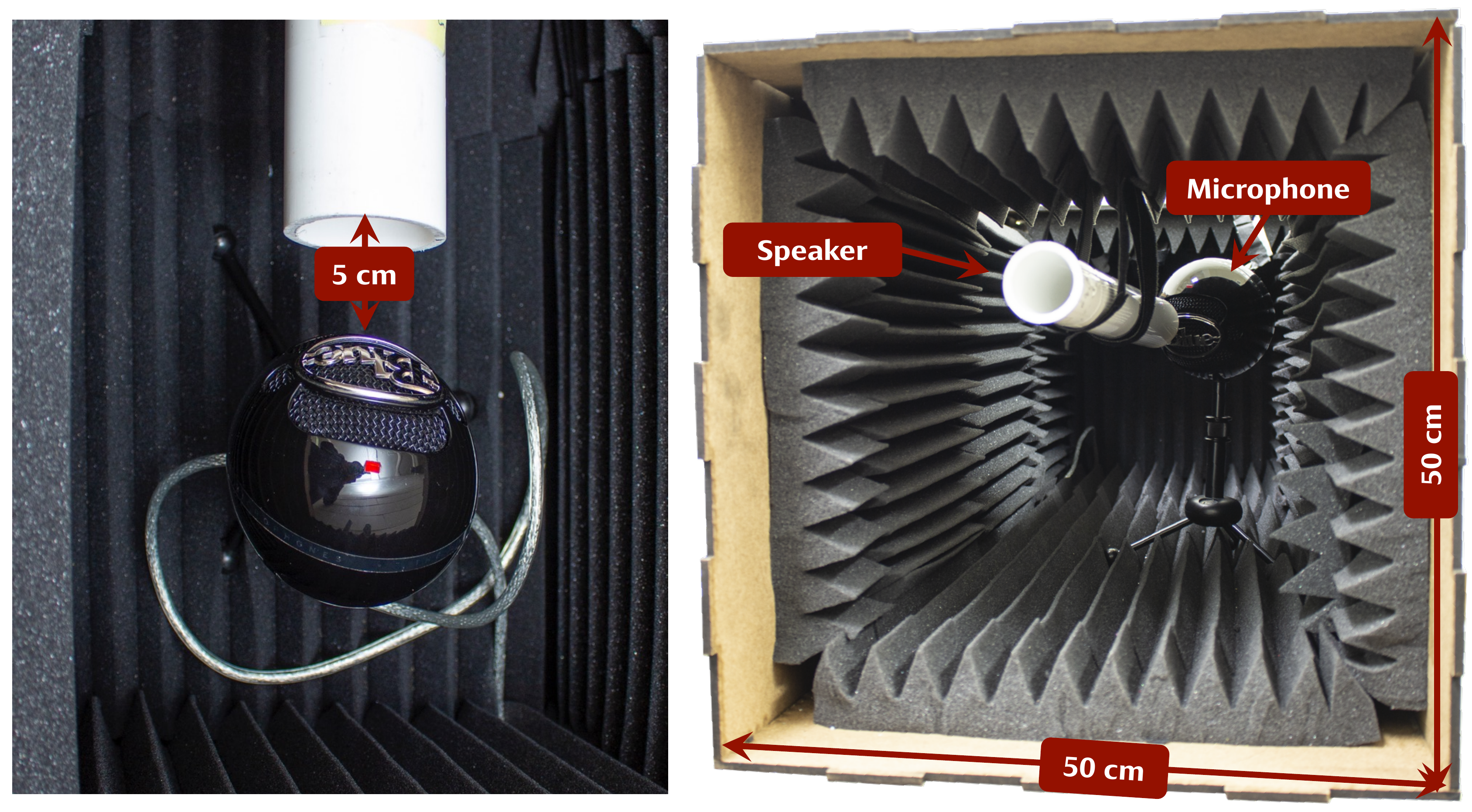}}
    \caption{The recording setup: top view (left) and front view (right).}
    \label{fig:box_setup}
\end{figure}

We design and implement a measurement setup to conduct the attack over the air. Fig.~\ref{fig:box_setup} visualizes our setup which comprises tube(s),  a recording device, and the recording environment. 

\paragraph{Tubes.} We use two sets of tubes in this work. We conduct the single-tube experiments using six PolyVinyl Chloride (PVC) pipes purchased from a hardware store. Their dimensions are listed in Table~\ref{tab:box_matching}. \rev{For the two-tube structures, we 3D printed the tubes for a fine-grained control over the tubes radii which impacts the resonance frequency (Eqn.~\ref{eqn:2tubes_nonlinearEq}). We used Formlab's Form 2\footnote{\url{https://formlabs.com/3d-printers/form-2/}} printer and Black Resin\footnote{\url{https://formlabs.com/store/black-resin/}} material. We print the tubes with a 50 \textmu m  resolution for a smoother finish and a thickness of 2 mm, no support material was on the inside of the tube. The tubes are connected using High Density Fiberboard (HDF) rings at run time, for tube reusability, as shown in Fig.~\ref{fig:2tubes_structure} in the appendix. We constructed three two-tube structures whose dimensions are in Table~\ref{tab:box_matching_2tube}. For both sets, we select the tubes dimensions based on our observations from Sec.~\ref{sec:91_attack_scaled} experiment}.

\paragraph{Recording Environment.}
We conducted the experiment in a $8\times3.6\times3.6$ m lab space. We built an audio chamber to \mr{prevent interference of the tube's input and output sounds,} and isolate the experiment from the background noise and speech interference from adjacent rooms; this helps unify the acoustic environment throughout the experiments. 
The chamber is a wooden box lined with acoustic panels to absorb the noise and minimize reverberation. We attached floating suspension loops to the chamber's ceiling to hold the tube in the air as shown in Fig.~\ref{fig:box_setup}. Suspending the tube minimizes its surface mechanical vibrations. We used a Blue snowball microphone,\footnote{\url{https://www.bluemic.com/en-us/products/snowball/}} placed as Fig.~\ref{fig:box_setup}, to capture the tube output signal. The setup is inspired by the design of musical instruments measurement environments. We use a Google Pixel 2 phone as a digital speaker to play sound over-the-air.  The recording is controlled by a MacBook Pro laptop. We used \texttt{python-soundevice} library to automate the recordings\footnote{\url{https://python-sounddevice.readthedocs.io/en/0.4.4/}}.

\vspace{-3pt}
\section{\name's Evaluation Results}
\label{sec:eval}

We conduct the following experiments to answer the three questions from Sec.~\ref{sec:exp_setup} in detail. 

\subsection{Impersonation Attack at Scale}
\label{sec:91_attack_scaled}

First, we test \name's impersonation attack feasibility on the full test set to address the first evaluation question. We run \name on the VoxCeleb \mr{(91 speakers)} test set, representing the adversarial speakers, and find the range of successful impersonation attacks and the corresponding set of adversarial tubes. In this experiment, we consider structures of N-tubes, where N $\leq$ 2. Hence, the resonating frequencies depends on three parameters (degrees of freedom): the tubes lengths $L_1$, $L_2$ and the tubes cross-sectional area ratio: $ratio_A=(d_2/d_1)^2$.

\mr{For each adversarial speaker, \name attempts to impersonate every enrolled speaker in the \ASI model; 7205 in SpeechBrain and 250 in X-Vector. \name searches for the BPF filters parameters that trick the \ASI into identifying the adversarial utterance $\operatorname{F_{tube}}(s, p)$ as the target victim speaker, $y_t$, using the DE algorithm~\ref{alg:de_algo}.}

\begin{figure}[t]
    \centering
    \scalebox{0.8}{
  \includegraphics[width=0.9\columnwidth]{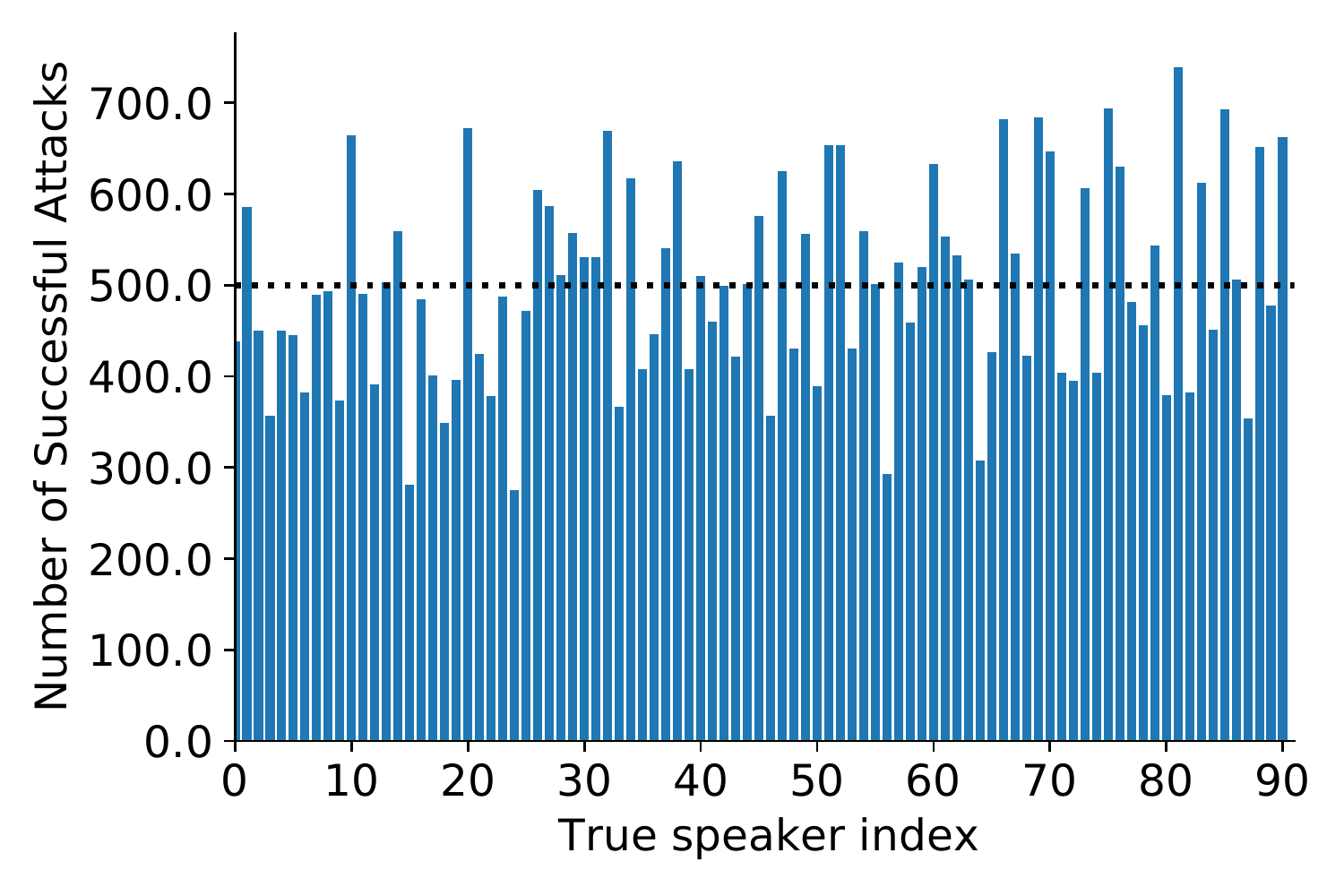}}
    \caption{\mr{Successful impersonation attacks (out of 7205) on SpeechBrain model for each adversarial speaker from VoxCeleb. Dotted line shows the average number of successful attacks per speaker.}}
    \label{fig:2tube_target_ids_HF}
\end{figure}

\mr{Fig.~\ref{fig:2tube_target_ids_HF} shows the number of target ids from SpeechBrain that an attacker could impersonate using Mystique \textit{theoretically}; i.e., the number of target ids where Mystique successfully finds a tube configuration that fools the model for each attacker. Since real-world requirements constrain the search and \name has little degrees of freedom, the algorithm might not find a tube for each source-target pair.} Fig.~\ref{fig:2tube_target_ids_xv} in appendix shows the same for the x-vector model. 
As the figure shows, \mr{by optimizing the tube dimensions, \name can successfully impersonate a wide range of victim speakers. Specifically, a speaker can impersonate 500 (out of 7205) target speakers on average on SpeechBrain model and 137 (out of 250) on x-vector model}. Recall that the models are initially 100\% accurate on the selected evaluation dataset. Hence, this experiment shows that \name is capable of forming an adversarial impersonation attack on \ASI models. Next, we analyze \mr{ the adversarial tube (BPF) parameters and the demographic distribution of the predictions to interpret how the attack works.} We report three findings.

First, the attack is most effective when $f_0$ lies in the frequency range $f_0 \leq 400 $ Hz with a high quality factor $Q_0 \geq 50$ as shown in Fig.~\ref{fig:fQ_histogram}. This observation matches our intuitions from Sec.~\ref{sec:intuition}; the significant $f_0$ range falls within the typical human pitch range. An adult woman pitch range is 165 to 260 Hz on average, and an adult man's is 85 to 155 Hz. Moreover, low frequency speech range carries more information than the higher frequency range~\cite{mary2018searching}. Hence, this range of $f_0$ will have a stronger impact on the pitch, the significant spectrum, and  the model prediction. Also, a high quality factor means a sharper filter; fine-grained selection.

Second, \mr{\name is 80\% more successful on impersonating same-sex targets than cross-sex.} Fig.~\ref{fig:CM_gender} in appendix shows the prediction confusion matrix split by the attacker-victim speakers sex. The figures show that the cross-sex speakers submatrix is sparser than that of the same-sex.
    
Third, \mr{we find that \name impersonates different victims when optimizing for different utterances of the same speaker (attacker). Hence, the attack is not utterance (text) independent. We attribute this observation to two reasons: (1) \ASI models are not perfect in separating the linguistic content and voice biometrics; the model prediction varies with the spoken utterance, (2) the attack's pitch shift and voice transformation is the resultant of \name's transformation applied on the spoken utterance original spectral content.}

\vspace{-2pt}

\subsection{Over-the-air Attack}

\label{sec:over-the-air}
We validate \name's impersonation attack over-the-air using our physical setup in Fig.~\ref{fig:box_setup} to answer the second evaluation question. We conduct this experiment on VoxCeleb as a standard dataset for \ASI---Sec.~\ref{sec:physical_vox}, and also on live speech from our user study participants---Sec.~\ref{sec:user-study}.

\subsubsection{Standard Dataset Evaluation}
\label{sec:physical_vox}

\begin{table*}[ht]
\small
    \centering
\adjustbox{max width=.8\textwidth}{%
    \begin{tabular}{c cc cc ccc ccc}
    \toprule
    {\multirow{2}{*}{\textbf{Tube}}} & \multicolumn{2}{c}{\textbf{Tube Dimensions}}  & \multicolumn{2}{c}{\textbf{Resonance Parameters}} & \multicolumn{3}{c}{\textbf{X-Vector False Predictions}} & \multicolumn{3}{c}{\textbf{SpeechBrain False Predictions}} \\
     &$L$ (cm) & $d$ (cm) & $f_0$ (Hz) & $Q_0$ & Real & Filter & Match
    & Real & Filter & Match \\
    \midrule
    1 & 40.6 & 3.45  & 402.16 & 58 & 158 & 141 & 64 (45.40\%)
     & 158 & 238 & 111 (46.64\%)\\
    
    2 & 61.3 & 4 & 270.70 &  68 & 123 & 194 & 75 (38.66\%)
     & 134 & 255 & 106 (41.57\%)\\
    
    3 & 87 & 5.2 & 191.48 &  77 & 202 & 242 & 101 (41.74\%)
    & 198 & 308 & 141 (45.8\%)\\
    
    4 & 99.4 & 3.45 &  170.89 &  64 & 325 & 174 & 77 (44.25\%) 
     & 220 & 200 & 121 (60.5\%) \\
    
    5 & 120.3 & 5.2 & 140.20 & 79 & 190 & 167 & 95 (56.89\%)
     & 210 & 351 & 146 (41.6\%)\\
    6 & 154 & 5.2 & 110.36 & 76 & 176 & 108 & 63 (58.34\%)
    & 179 & 185 & 114 (61.62\%)\\

    \bottomrule
    \end{tabular}
}
    \caption{Evaluation of \name over-the-air for 40 speakers $\times$ 20 utterances: 800 total inferences. \textit{Real}: \# successful attacks of the real tube, \textit{Filter}: \# successful attacks of the corresponding filter model, \textit{Match}: the number (percentage) of matched attacks between filter and real tube.}
    \label{tab:box_matching}
\end{table*}

\begin{table}[t]
\small
    \centering
\adjustbox{max width=\columnwidth}{%
    \begin{tabular}{c cccc c c}
    \toprule
    {\multirow{2}{*}{\textbf{Tube}}} & \multicolumn{4}{c}{\textbf{Tube Parameters} (cm)}  &{\multirow{2}{*}{\textbf{$f_0$} (Hz)}} & {\multirow{2}{*}{\textbf{Attack Success Rate}}} \\
     &$L_1$ & $d_1$ & $L_2$ & $d_2$ &  &  \\
    \midrule
    7 & 9.53 & 2.1 & 10 & 1 & 853.1 & 66.6\%\\
    8 & 11.44 & 0.98 & 8.9 & 3.4 & 901.55 & 50\%\\
    9 & 14.53 & 2.1 & 10 & 1 & 600.4 & 100\%\\
    \bottomrule
    \end{tabular}
}
    \caption{Two-tube structures $f_0$ and  attack success rate over-the-air.}
    \label{tab:box_matching_2tube}
\end{table}

 Because of the physical resources (mainly run-time) limitations, we select a subset of the evaluation speakers to form the adversarial speakers set. We also select a subset of the possible tube dimensions to run the over-the-air attack.
Specifically, we randomly select 40 speakers, 20 males and 20 females, out of the 91 speakers dataset. There are 20 utterances for each speaker; a total of 800 four-second long utterances. The subset is balanced and representative of the full dataset. For the single-tube setting, we select 6 random tubes of various dimensions\rev{, listed in Table~\ref{tab:box_matching},} which have $f_0, Q_0$ in the most significant range---Fig.~\ref{fig:fQ_histogram}. We purchase them from the hardware store. While for the two-tube setting, we build three structures of 3D printed tubes \rev{as described in Sec.~\ref{sec:physical_setup}; their parameters are listed in Table~\ref{tab:box_matching_2tube}.}

We use the Pixel phone to simulate the speaker and play the VoxCeleb utterances over-the-air for all tubes. We record the tube output sound using the physical setup. We place the speaker on a separate tripod to allow acoustic propagation only through the air; \ie, no sound is transmitted to the microphone via vibrations through the recording table. We allow a 3 sec silence between consecutive utterances. We repeat the recordings 6 times to account for any environmental variations \mr{and to evaluate the attack reliability and consistency.}

\paragraph{Single-Tube.} Table~\ref{tab:box_matching} shows the number of \mr{successful attacks (impersonated targets) per tube} and compares it to the successful attacks using the filter model. First, ``Real'' columns (6 and 9) report the number of successful attacks of the 40 speakers using the real tubes. Each speaker can impersonate up to 5 speakers identities on average using an individual tube, \mr{depending on the attacker's spoken utterance. As discussed in Sec.~\ref{sec:91_attack_scaled}, we found that different utterances sometimes lead to different impersonated victims per attacker-tube pair. Second, ``Filter'' columns (7 and 10) show the number of successful attacks using each tube's BPF model. } The filter's successful attacks are on the same magnitude as the real tube. 
Finally, the ``Match'' columns (8 and 11) show the matching rate between the real and simulated tubes attacked identities. The match rate ranges from 38.7\% to 61.62\%, 48\% on average. Hence, Table~\ref{tab:box_matching} confirms that speaking through a tube forms a real and effective attack on the \ASI task\mr{, and the linear BPF model (Eqn.~\ref{eqn:outputSignal},~\ref{eqn:fQi}) is a reasonable approximation of the resonance effect}. 
A more accurate model is to use wave simulation engines at the expense of increased computation complexity. 

\mr{Finally, we assess the attack's reliability over multiple trials. We measure the model's predictions consistency rate---defined as the percentage of consistent predictions across six runs. Table~\ref{tab:consistency} in appendix shows the consistency rate per tube, on average 84\% of the predictions are consistent over six runs.}

\rev{ 
\paragraph{Two-Tube.} Similarly, Table~\ref{tab:box_matching_2tube} shows \name's performance over-the-air using the two-tube configurations. The success rate is the percentage of matched successful attack between Filter and Real tubes impersonated identities. \name's targeted attack succeeds more than 50\% of the time. }

\subsubsection{Live Impersonation Attack}
\label{sec:user-study}

\begin{figure}[t]
    \centering
    \scalebox{0.75}{
  \includegraphics[width=0.9\columnwidth]{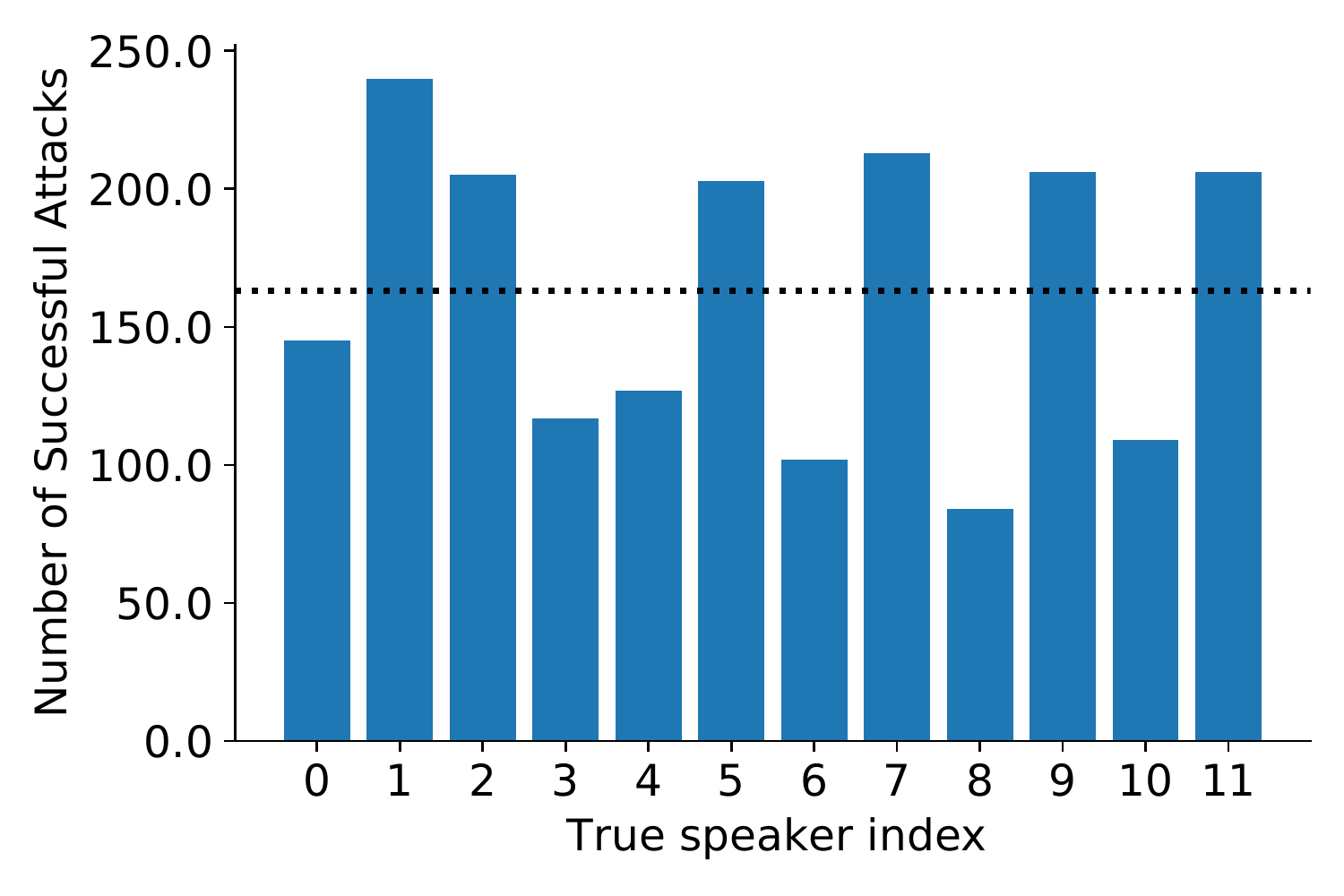}}
    \caption{Number of successful attacks of the study participants recordings on SpeechBrain. The dotted line shows the average number per true speaker.}
    \label{fig:1tube_target_ids_HF_userstudy}
\end{figure}

We run \name on 14 participants natural recordings, 50 utterances each, and find the set of \mr{theoretically} successful attacks (impersonated identities) per participant. Fig.~\ref{fig:1tube_target_ids_HF_userstudy} shows the number of successful attacks on the SpeechBrain model. Fig.~\ref{fig:1tube_target_ids_xv_userstudy} in the Appendix shows the same for the x-vector model. An arbitrary speaker can impersonate 163 (117 for x-vector) target identities on average \mr{using a single-tube.} 

Next, we ask the participants to speak the same 50 utterances through three of our tubes. \mr{We evaluate the recordings on the \ASI models and compare them to the BPF predictions. Table~\ref{tab:user_match} reports the percentage of \name's BPF impersonation attacks that also succeeded over-the-air in the live recording of each participant. The average success rate ranges from 34.84\% to 78.25\%, showing that \name reliably launches over-the-air attacks. This result is significant---live human speech varies between recording sessions unlike \eg~VoxCeleb experiment with fixed recordings.}

\mr{Moreover, we explore \name's personalization} by fine-tuning the filter parameters to each participant's voice characteristics. Applying a voice envelope calibration to the filter gain increases \name's success rate, for most participants, up to 10\%. \mr{However, it drops for a few participants, as shown in column 6 of Table~\ref{tab:user_match}. Thus, personalization is one way to further optimize \name, which we leave to future work.} Additionally, we observe the same skew in the speaker's sex for successful attacks as in VoxCeleb (Fig.~\ref{fig:CM_gender}), where the cross-sex submatrix is sparse.

\begin{table}[t]
\small
    \centering
\adjustbox{max width=\columnwidth}{%
    \begin{tabular}{c c cccc c c}
    \toprule
    \textbf{ID} & \textbf{Gender} & \textbf{Tube3} & \textbf{Tube4} & \textbf{Tube6} & \textbf{Avg} & \textbf{\mr{Avg$_{Cal}$}} & \textbf{\mr{Human}}\\
    \midrule
    0 & F & 50.0 & 50.0 & 66.67  & 55.56 &  \textbf{65.0} & 0/5\\  
    1 & M & 58.82 & 81.82 &  57.14 &  65.93 & 43.02 & 0/40\\ 
    2 & M & 66.67 & 72.73 & 77.78 & 72.40 & \textbf{72.58} & 0/20\\ 
    3 & F & 63.64 & 83.33 & 75.0 & 73.99 & \textbf{78.7} & 0/20 \\ 
    4 & F & 66.67 & 58.33 & 71.43 & 65.48 & \textbf{73.15} & 0/20 \\ 
    5 & M & 50.0 & 42.86 & 55.56 & 49.47 & 42.29 & 1/20 (5\%)\\ 
    6 & M & 46.15 &  54.55 &  80.0 & 60.23 & \textbf{60.71} & 6/25 (24\%)\\
    7 & F & 66.67 & 77.78 & 80.0 &  74.81 & 62.22 & $-$ \\
    8 & M & 43.75 & 42.86 & 54.55 &  47.05 & \textbf{52.06} & 0/10\\
    9 & M  & 50.0 &  60.0 &  50.0 & 53.33 & 41.6 & 1/10 (10\%)\\ 
    10 & F & 66.67 & 62.5 & 80.0 &  69.72 & \textbf{75.0} & 5/20 (25\%)\\
    11 & M & 50.0 & 61.54 & 72.73 & 61.42 & \textbf{69.17} & $-$ \\
    12 & M & 10.0 & 54.55 & 40.0 & 34.84 & \textbf{35.56} & 0/15 \\
    13 & F & 80 & 71.42 & 83.33 & 78.25 & 69.44 & 1/20 (5\%)\\
    \bottomrule
    \end{tabular}
}
    \caption{User study participants percentage (\%) of successful over-the-air impersonation attacks with and without \name. \mr{\textbf{bold} values are enhanced by personalized calibration.}}
    \label{tab:user_match}
\end{table}

\mr{Finally, we evaluate the participants impersonation capabilities without using any tubes as a baseline for \name's performance. The last column in Table~\ref{tab:user_match} shows the number of times the participant was able to impersonate a target by the total number of trials, where for each target the participant performs 5 impersonation trials. Note that some participants were not willing to impersonate more than one target, thus the total number of trials is not the same for all of them. This study shows that most participants were not able to reliably impersonate a target speaker, where the average success rate is only 6.22\%. Specifically, 7 participants did not succeed in any trials, 3 participants were able to impersonate one target one time and failed at the 4 other attempts for the same target, and only 2 participants could succeed more than once. We noticed they could capture the accent and pitch of the target. Participant 6 impersonated 3 (out of 5) targets for (3, 2, 1) trials for the same target, while participant 10 successfully impersonated 2 targets for (2, 3) times. Yet, \name significantly outperforms the strongest baseline; it impersonates 100+ victims with success rate of up to 78.7\%.}

\vspace{-3pt}

\subsection{\name's Robustness} %
\label{sec:attack_reliability}

\mr{We study different strategies to detect samples from \name, which include: comparing prediction confidence, human-based analysis, and state-of-the-art spoofing detection.

\subsubsection{What is the \ASI model confidence on \name?}
\label{sec:prediction_confidence}
The \ASI model outputs the class (speaker id) with the highest prediction score. Here, we analyze the model's confidence of the predicted class, using the softmax score as a proxy. 

Fig.~\ref{fig:score_predictions_top2} shows the distribution of the model's top two classes confidence scores in case of clean (benign) and \name (adversarial) samples. The figure shows that the model is less confident of its top-1 class prediction on \name's samples; i.e., the gap between the top-2 scores decreases. This finding arises from \name's samples being out-of-distribution (OOD) samples with respect to the model's training data. Hence, this analysis suggests that \name's threat can be weakened if the \ASI model is trained to reject samples of which it is not highly confident~\cite{rabanser2022selective}.

\begin{figure}[t]
    \centering
    \scalebox{0.65}{
  \includegraphics[width=\columnwidth]{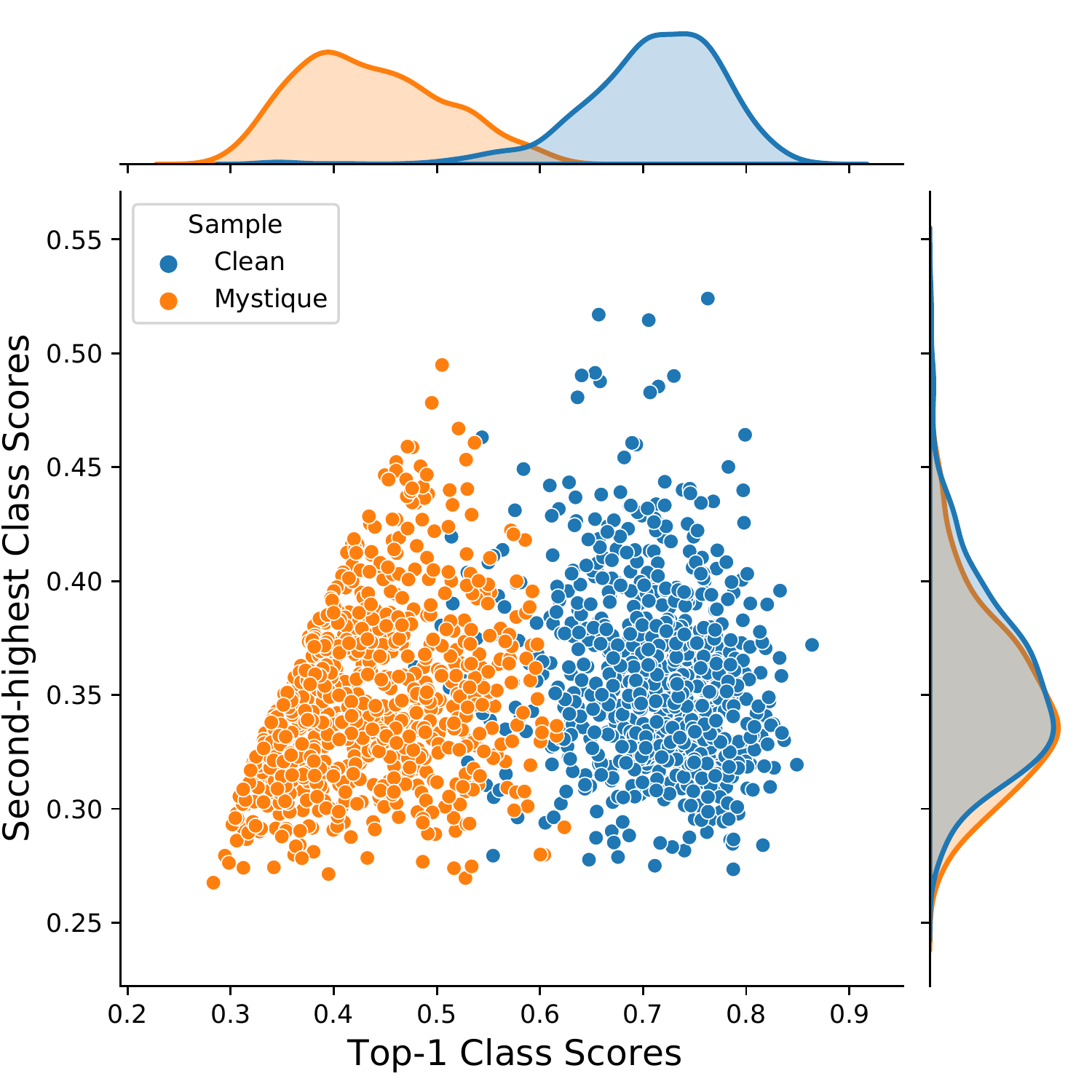}}
    \caption{\mr{The distribution of BrainSpeech softmax scores for the top two classes on VoxCeleb clean and adversarial samples.}}
    \label{fig:score_predictions_top2}
\end{figure}

\subsubsection{How similar is \name to the victim's voice?}
\label{sec:attacker-vicitm_similarity}
Our second detection strategy assesses \name's spoofed speech similarity to the victim's speech. We analyze the attack-victim speech pairs that are successful over-the-air. Specifically, we visualize the attack in the problem space (audio) and evaluate the similarity in the embedding space.

Fig.~\ref{fig:attack-victim-signals} in Appendix~\ref{app:more_results} shows a sample of the attacker and victim waveforms and spectral content. The samples are not visually similar and do not exhibit high cross correlation. We attribute this result to speech being composed of two entangled characteristics: the speaker's voice and the linguistic content~\cite{yuan2021improving,ahmed2020preech}. The attack-victim samples are of different linguistic content. Note that VoxCeleb is composed of Youtube recordings of celebrities, and there is no transcript or one-to-one mapping of the linguistic content among different speakers. 
However, the \ASI models learn a representation of the speech utterances that are presumably based on voice characteristics and invariant to the linguistic content.

Next, we analyze the \ASI model embedding space similarity of the attack-victim pairs. Table~\ref{tab:cosine_sim_embedings} shows that the embeddings of attack and victim sample pairs exhibit high cosine similarity, compared with randomly selected non-victim speakers. Thus, regardless of the linguistic content, \name applies a transformation on the speech utterance that maps its embedding (voiceprint) towards the victim's voiceprint.

\begin{table}[t]
\small
    \centering
\adjustbox{max width=\columnwidth}{%
    \begin{tabular}{l cccccc }
    \toprule
     \multirow{2}{*}{\textbf{Cosine Similarity}} & \multicolumn{6}{c}{\textbf{Tubes}}\\
     & 1 & 2&3&4&5&6\\
    \midrule
    Attack-Victim & 0.39 & 0.38 & 0.40 & 0.43 & 0.38 & 0.34\\
    Attack-Non victim & 0.07 & 0.07 & 0.08 & 0.08 & 0.08 & 0.08\\
    \bottomrule
    \end{tabular}
}
    \caption{\mr{Average cosine similarity score of the embeddings of \name's successful attack utterances and its victim speakers' utterances, compared to non-victim speakers similarity scores.}}
    \label{tab:cosine_sim_embedings}
\end{table}
\vspace{-3pt}

\subsubsection{Does \name confuse humans as well?}
\label{sec:prolific_study_similarity}
Here, we investigate the similarity from humans point of view.
Although \name is designed to attack \ASI ML models by physically manipulating the  spectral content of speech, we are curious whether it also confuses humans. To answer this question, we recruit participants to listen to two audio recordings and decide whether they belong to the same speaker, and also rate the audio quality as natural or unnatural. The study is approved by IRB and is conducted on the Prolific platform.

\paragraph{Study design.} We recruited 151 participants, each compensated \$1.4 for their effort,
with an average completion time of 6 minutes. 
Each participant listens to 10 pairs of audio recordings from VoxCeleb; 3 pairs from each of the following cases: (a) the two recordings are clean and belong to the same speaker, (b) the two recordings are clean and belong to two different speakers of the same sex, and (c) one recording is generated by \name (attacker using a tube), while the other recording is of the corresponding victim's voice. The tenth pair is an attention check with two identical clean recordings. For each pair of recordings, we ask the participants two questions: (1) ``\textit{do they belong to the same speaker?},'' and (2) ``\textit{how natural does the recording sound?}'' on a 3-point Likert scale. We discard any responses that did not answer ``same speaker'' for the attention checker.

\paragraph{Results.} Fig.~\ref{fig:prolific_responses} shows the distribution of responses.
Fig.~\ref{similarity} shows that \name generated successful attacks on humans perception 16\% of the time, and was able to confuse them 12\% of the time. This result is interesting given that \name is not optimized to trick humans. The study also shows that the participants could distinguish different speakers voices with high probability (89\%). However, they were confused on the ``Same'' speaker recordings. Here, 44\% were labeled as same (different) speaker; \ie \: not significantly better than random guessing. This result, supported by previous studies~\cite{lavan2016impaired, winters2008identification}, confirms that the voice identity perception can be challenging for humans, specially of unfamiliar speakers. Finally, Fig.~\ref{naturalness} shows that 63\% of \name's samples sound unnatural to the participants, yet 14\% sound natural. These results show how ML models perceive speech differently than humans, creating the gap which \name and other attacks exploit.

\begin{figure}
  \centering
     \begin{subfigure}[c]{0.5\columnwidth}
         \centering
         \includegraphics[width=\textwidth]{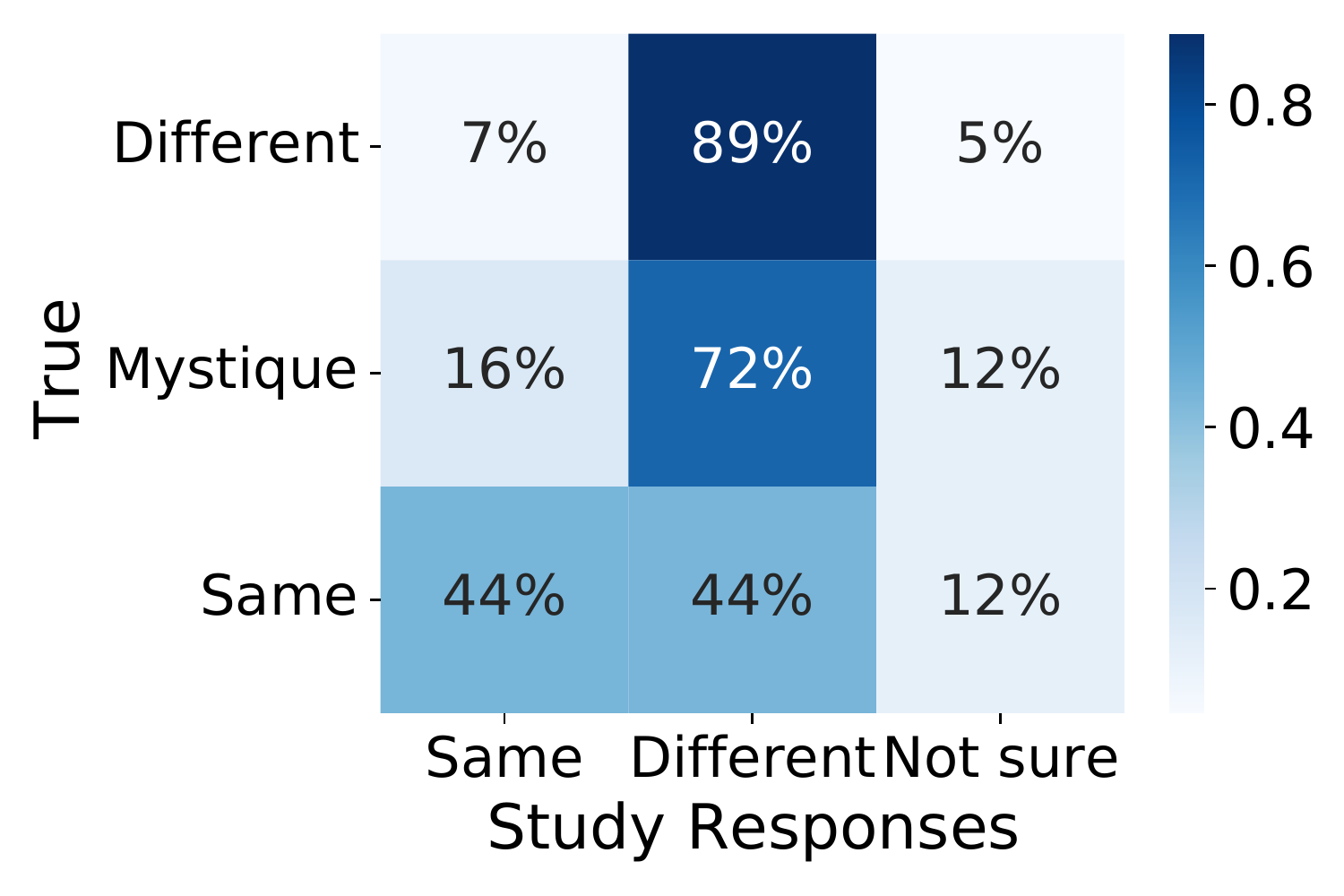}
         \caption{Similarity}
         \label{similarity}
     \end{subfigure}
     \begin{subfigure}[c]{0.48\columnwidth}
         \centering
         \includegraphics[width=\textwidth]{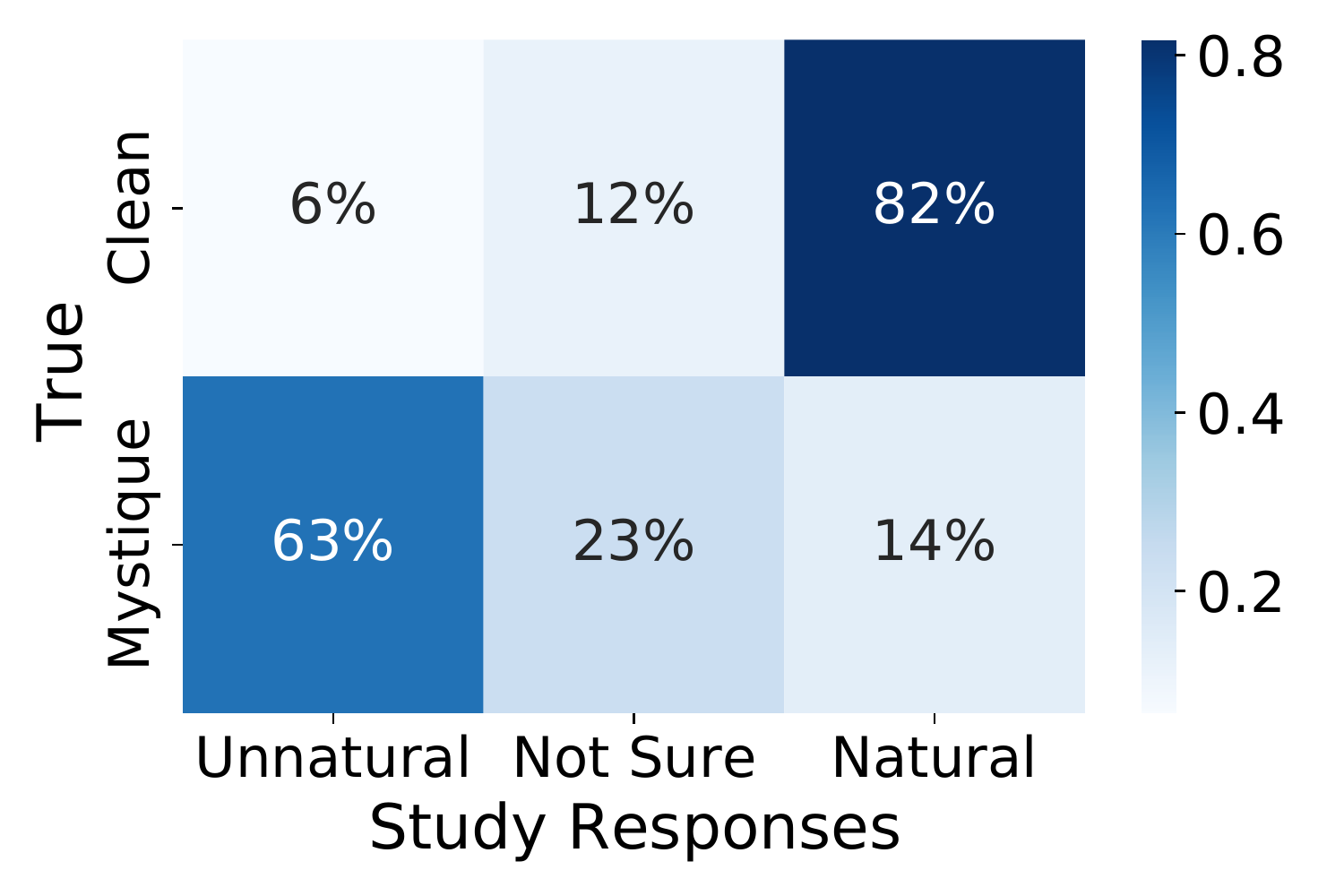}
         \caption{Quality}
         \label{naturalness}
     \end{subfigure}
     \caption{The confusion matrix of the user study responses on the audio recording similarity and quality evaluation. } 
     \label{fig:prolific_responses}
\end{figure}

\subsubsection{Spoofing Detection}
\label{sec:spoof_detection}

\begin{table}[t]
\small
    \centering
\adjustbox{max width=\columnwidth}{%
    \begin{tabular}{l cc cc cc }
    \toprule
    \multirow{2}{*}{\textbf{Model}} & \multicolumn{2}{c}{\textbf{Pre-trained}} & \multicolumn{2}{c}{\textbf{VoxCeleb}} & \multicolumn{2}{c}{\textbf{VoxCeleb$_{3\:tubes}$}} \\
     & {EER} & {FAR$_0$} & {EER} & {FAR$_0$} & {EER} & {FAR$_0$}\\
    \midrule
    \rev{LA-LCNN} & 29.59 & 76.67 & 6.25 & 19.21 & 8.54 & 22.30\\
    \rev{PA-LCNN} & 31.33 & 98.9 & 7.87 & 33.94 & 14.54 & 67.81\\
    Void-SVM & 62.12 & 97.70 & 36.70 & 91.64 & 48.3 & 96.12 \\
    Void-DNN & 35.39 & 91.88 & 26.12 & 86.61 & 39.1 & 92.55 \\
    Void-LCNN & 32.91 & 93 & 27.03 & 91 & 27.64 & 92.55 \\
    \bottomrule
    \end{tabular}
}
    \caption{\mr{Evaluation of five spoofing detectors on the user study recordings for: (1) pre-trained detectors, (2) fine-tuned on VoxCeleb clean and tube recordings, (3) fine-tuned on VoxCeleb \textbf{without} the three tubes used in the user study. Note: FAR$_0$ = FAR at FRR=0\%.}}
    \label{tab:spoofing_detectors}
\end{table}

Finally, we assess \name's robustness against a set of \rev{five} representative defenses and liveness detectors as detailed in Sec.~\ref{sec:models_implemention}. For the five evaluated models, we report the equal error rate (EER) which is the model's error rate when the false acceptance rate (FAR) and false rejection rate (FRR) are equal. A lower EER means a more accurate detector. We also report the FAR at 0\% FRR, which sets the detector's operating threshold to correctly classify all live samples. 

\rev{Table}~\ref{tab:spoofing_detectors} reports the detectors performance on the  recordings of 12 participants from the user study (Sec.~\ref{sec:user-study}). We label the no-tube speech as \textit{bonafide} \rev{(live)} and the tubes recordings as \textit{spoofed}. The \rev{``Pretrained''} column reports the performance of the pretrained models on ASVspoof dataset. All \rev{evaluated} models are unable to distinguish the tube and no-tube samples. The best performing model (\rev{LA-LCNN}) labels 76\% of the tube samples as bonafide. Note that these models perform very well on the synthetic data from their original papers. For example, \rev{LA-LCNN}'s EER = 9.26\% on the ASVspoof test set~\cite{yamagishi2021asvspoof} and Void's EER < 12\% on the curated synthetic dataset by Ahmed et al.~\cite{ahmed2020void} and Wenger et al.~\cite{wenger2021hello}. Yet, spoofing detectors do not generalize to \name's samples. 

Next, we retrain the models on our VoxCeleb clean and tube recordings as the bonafide and spoofed samples respectively. The goal is to introduce \name's resonance effect to the model and label it as spoofed. \rev{The ``VoxCeleb'' column} in Table \ref{tab:spoofing_detectors} shows that the EER and FAR$_0$ dropped for all models, especially \rev{LA-LCNN} with only $19\%$ of the tube samples labeled as bonafide. Then, we retrain the models again on VoxCeleb but without the recordings from the tubes (3, 4, 6) that we use in the user study. \rev{The last column ``VoxCeleb$_{3tubes}$''} EER and FAR$_0$ values increase by $2\%$ to $85\%$ relative to \rev{the VoxCeleb column}. When the exact tubes used by the participants \rev{in the test set} are not part of the training data, the performance drops significantly. Thus these models overfit to their training distributions.  Previous work~\cite{pianese2022deepfake, muller2022does,  borzi2022synthetic} has reported the same observation; spoofing detectors hardly generalize to unseen transformations in the training data, which questions the security of voice-based authentication.}

\section{Discussion}
\label{sec:discussion}

\paragraph{Defenses.} Having established a major vulnerability in spoofing detection systems leads to a question on how one stops such attacks. \mr{We show that defenses trained on samples of the resonance transform can detect \name and limit its effectiveness.} However, it is not clear whether such a defense approach is reliable, or even desirable. An attacker can simply use objects with different filter profile to render the defense unsuccessful; the defender cannot predict what filter the attacker would deploy. A better defense would have to incorporate properties of the medium \mr{and other modalities to rely on multiple factors}, not just the speakers features.

\paragraph{Reproducibility.} From formulating the original idea to completing the experiments of this paper, this work took around a year. We make a note of the things that slowed us down significantly and required non-trivial debugging. First, the use of Bluetooth or Wifi operated devices introduces significant problems because of occasional variable lag and interference. Second, during the theoretical and practical matching, it is important to isolate the setup as much as possible. In our case, matching $f_0$ and $Q$ \mr{without the acoustic chamber} was extremely challenging. Third, distance to the microphone and its' directionality matters---nothing should be blocking the opening of the tube, as otherwise it leads to additional echo and changes the filter \mr{as reported in the measurements literature~\cite{end_correction}.} Fourth, experiments ran on different days lead to different results, because of a change in speed of sound with temperature and humidity -- its best to conduct hardware calibration and the evaluation on the same day. Finally, when producing tubes with a 3D printer, the material on the inside of the tube should be smooth. %

\paragraph{Limitations.} Despite highlighting a flaw in current defenses design, there are a number of limitations in the current evaluation. First, we only considered simple tube structures, restricting the range of possible adversarial transformations. 
\rev{Second, we run the attack in a static recording environment, limiting its deployment in more practical situations where the adversary can be visually-observed or has partial control over the acoustic environment or experiences acoustic effects such as noise and interference.} Third, we evaluated a small number of speakers and utterances, potentially underrating the overall attack performance. \rev{Fourth, the resonance effect sounds unnatural to humans, other transforms should be explored to have a more subtle impression on human listeners. Finally, \name needs access to the \ASI model scores to perform the DE algorithm. However, \name can omit this requirement and perform an exhaustive search over all possible tube parameters, at the expense of the time complexity.}

\paragraph{Future Work.} 
We provide some directions to address \name's limitations. First, physical effects such as natural sounds and acoustic meta-materials \rev{should be explored to provide higher degrees of freedom and a less susceptible attack. Second, our evaluations suggest that the linguistic content can be optimized per each attacker-victim pair. Moreover, Table~\ref{tab:user_match} suggests that attack personalization can boost its success rate. 
Third, \name can be made model-independent by performing the optimization on the estimated pitch as a proxy of the \ASI model's decision as explained in Sec.~\ref{sec:intuition},~\ref{sec:pitch_shift}.}

\section{Related Work}
\label{sec:related}

The literature on computer-based voice authentication is vast, and dates back to at least 1960s~\cite{kersta1962voiceprint}. 

\paragraph{Attacks on ASI.} We start by describing the four most common attacks: (1) speech synthesis, (2) voice conversion, (3) replay attacks and (4) adversarial examples. In \textit{speech synthesis}, an adversary trains a speech synthesis model on samples recorded from the victim speaker. The adversary uses this model to convert text into speech in the victim’s voice~\cite{taylor2009text,oord2016wavenet,wang2017tacotron}. Alternatively, voice conversion converts spoken utterances into the victim’s voice~\cite{muda2010voice,sisman2018adaptivewavenet,yuan2021improving}. In \textit{replay} attacks, the adversary records the speaker’s voice and replays the recorded speech~\cite{Lindberg1999vulnerabilityin}. Finally, many modern ML-based \ASI models inherit the vulnerability to adversarial examples using standard  gradient-based attacks~\cite{liu2019adversarial,kassis2021practical,chen2021spoofingspeaker}.

\paragraph{Defenses against Acoustic Attacks.}
What these attacks have in common is that the adversarially-generated sample would need to be generated, and transmitted digitally and reproduced through a (digital) speaker. Defense mechanisms, therefore, include (1) detecting the electronic footprint of the digital speaker (known as spoofing detection), or (2) verifying that the speaker is a live human.

Spoofing detection relies on patterns extracted from the acoustic signal to classify it as a legitimate or fake sample. Chen et al.~\cite{chen2017youcanhear}~used a smartphone's magnetometer to detect the use of a loudspeaker.  Blue et al.~\cite{blue2018helloisitme} tell electronic and human speakers apart by analyzing individual frequency components of a given speech sample. 
Yan et al.~\cite{chen2019thecatcher} calibrated individual speakers in the near field of the speakers to tell humans and electronic speakers apart. 

Second, liveness detection leverages other sensing modalities such as visual, acoustic and EM signals to determine the liveness of the acoustic signal. Meng et al.~\cite{meng2018wivo}~used an active radar to project a wave onto the face of the speaker and then detect shifts introduced to it from facial movement.
Zhang et al.~\cite{zhang2017hearingyourvoice} analyzed hand movement to detect live speech by turning a smartphone into an active sonar.

Finally, there exists a class of defenses that restrict the attack surface by reducing attacker capabilities. Zhang et al.~\cite{zhang2016voicelive} used individual recordings from a stereo microphone to calculate time difference of arrival to detect replay attacks. Blue et al.~\cite{blue20182ma}~used two microphones to restrict the adversary to a 30 degree cone and protect against hidden and replay commands. Wang et al.~\cite{wang2019defeatinghiddelaudio} used correlates from a motion sensor to detect and reject hidden voice commands.

\paragraph{Physical Adversarial Examples.} \textit{Physical} adversarial examples are common in the vision domain, but have not been produced for acoustic tasks. Example adversarial objects include eyeware~\cite{chen2017targeted,sharif2019ageneral}, tshirts~\cite{wu2020making,xu2020adversarial}, headwear~\cite{Komkov2021advhat,zhou2018invisible} and patches~\cite{thys2019fooling}. Although these objects were re-created in the real world, there is an important distinction here. 
These objects all apply perturbations that were initially designed for the digital space and then retrofitted with sophisticated machinery such at printers to realize them in the physical domain. 
Our attacks, on the other hand, 
directly restrict the search space of perturbations to those that can be easily realized physically.
Most importantly, our attacks target a different property of the physical world---we use the environment to shape the signal, rather than exploit errors in the ML model.

\section{Conclusion}

We demonstrate that a human adversary can reliably manipulate voice-based identification systems using physical tubes, \textit{without access to the victim's speech}. Our attacks highlight acoustic intricacies that were largely ignored by prior literature, namely, the acoustic environment. %
Current defenses assume that the adversary is non-human and focus on verifying this assumption. Our human-produced attacks show that this assumption does not hold in the first place.
 In this paper, we demonstrate that subjective nature of speech can be exploited to jeopardize the security of a critical system. 
 \rev{Concretely, a fundamental question to consider in speaker identification is whether a person's identity can be accurately established despite the transformation of their voice.}

\section*{Acknowledgment}
This work was supported by DARPA (through the GARD program), the Wisconsin Alumni Research Foundation, the NSF through awards: CNS-1838733 and CNS-2003129, CIFAR (through a Canada CIFAR AI Chair), NSERC (under the Discovery Program and COHESA strategic research network), a gift from Intel, and a gift from NVIDIA. We also thank the Vector Institute's sponsors. Finally, we thank Bill Sethares, Andrew Allen, Nikunj Raghuvanshi, Hannes Gamper, Darija Halatova, and the reviewers for their fruitful discussions and recommendations. 

\bibliographystyle{abbrv}
\bibliography{ref}

\appendix
\section{Appendix}
\label{sec:appendix}

\subsection{Proof: Tubes Cause Pitch shift}
\label{sec:pitch_shift}
McAulay and Quatieri~\cite{pitch_estimation} use the peaks of the Short-time Fourier transform (STFT) of a time domain signal $s(t)$ to represent it as a sum of $L$ sine waves:
\begin{equation*}
\small
s[n] = \sum_{\ell=1}^{L}{A_\ell\exp[{j(n\omega_\ell) + \theta_\ell}]}.
\end{equation*}

The values of $A_\ell$, $\omega_\ell$, and $\theta_\ell$ represent the amplitudes, frequencies, and phases of the STFT peaks of the speech signal. Then, they find the value of $\omega_0$ which fits $s[n]$ to $\widetilde{s}[n,\omega_0]$ as: 
\begin{equation*}
\small
\tilde{s}[n,\omega_0] = \sum_{k=1}^{K(\omega_0)}{\widetilde{A}(k\omega)\exp[{j(nk\omega_0) + \phi_k}]},
\end{equation*}

where $\omega_0$ is the signal pitch, $K(\omega_0$) is the number of harmonics in the signal, $\widetilde{A}(k\omega)$ is the vocal tract envelope, and $\phi_k$ is the phase at each harmonic. Finally, the pitch is estimated by minimizing the mean squared error $\epsilon\left(\omega_{0}\right)=P_{s}-\rho\left(\omega_{0}\right)$, where $P_{s}$ is signal's power which is a constant. Therefore, we only need to minimize $-\rho(\omega_0)$, or equivalently:
\begin{align}
\label{eq:opt1}
\small
\centering
    &\max \quad \rho(\omega_0)
\end{align}
where 
\begin{equation}
\label{eq:rho}
\small
\rho(\omega_0) = \sum_{k=1}^{K(\omega_0)}{ \widetilde{A}(k\omega_0)}\left[\sum_{\ell=1}^{L}{A_\ell\left|\operatorname{sinc}\left(\omega_\ell-k\omega_0\right)\right|
                    -\frac{1}{2}\widetilde{A}\left(k\omega_0\right)}\right]. 
\end{equation}
As discussed in \Cref{sec:explain_resonance}, the tube results in a resonance effect, modeled as a set of bandpass filters at the resonance frequencies of the tubes. As such, some of the frequency components of $s(t)$ will be dampened. We represent this effect as $A_\ell=0$ for $\ell\in \mathcal{L}$ as well as their submultiples $\omega_0 \in [K(\omega_0)]$, where $\mathcal{L}$ represents the set of non-resonant frequencies:
\begin{align}
\label{eq:opt2}
\small
\centering
    &\max &\rho(\omega_0) &\\\nonumber
    &\text{s.t.} &A_\ell = 0 &\quad\quad\forall \ell \in \mathcal{L}, \forall \omega_0 \in [K(\omega_0)] 
\end{align}
Note that \eqref{eq:opt2} is a constrained version of \eqref{eq:opt1}. We can solve the latter by maximizing the Lagrangian:
\begin{equation}
    \label{eq:lagrange}
    p(\omega, \boldsymbol{\eta}) = \rho(\omega_0) - \sum_{k=1}^{K(\omega_0)}\sum_{\ell \in \mathcal{L}} \eta_{k\ell}A_\ell
\end{equation}
where the matrix $\boldsymbol{\eta} = [\eta_{k\ell}]_{K(\omega_0) \times |\mathcal{L}|}$ represents the Lagrange multipliers. Instead of directly maximizing \eqref{eq:lagrange} and finding $\boldsymbol{\eta}$, we re-write \eqref{eq:rho} separating the components in and outside of $\mathcal{L}$:
\begin{equation}
\label{eqn:final_obj}
\small
\rho(\omega_0) = \rho_f(\omega_0) +
 \sum_{k=1}^{K(\omega_0)}{ \widetilde{A}(k\omega_0)}\sum_{\ell\in\mathcal{L}}{A_\ell|\operatorname{sinc}(\omega_\ell-k\omega_0)|}.
\end{equation}
where
\begin{equation}
\label{eq:rho-f}
\small
 \rho_f(\omega_0) = \sum_{k=1}^{K(\omega_0)}{ \widetilde{A}(k\omega_0)}\Big[\sum_{\ell\not\in\mathcal{L}}{A_\ell|\operatorname{sinc}(\omega_\ell-k\omega)|-\frac{1}{2}\widetilde{A}(k\omega_0)}\Big],
\end{equation}
is the objective function for estimating the pitch of the filtered signal.
Next, substituting \eqref{eqn:final_obj} in \eqref{eq:lagrange}:
\begin{align}
\label{eq:lagrange-rewrite}
\small
p(\omega, \boldsymbol{\eta}) 
&= \rho_f(\omega_0) + \sum_{k=1}^{K(\omega_0)}\sum_{\ell \in \mathcal{L}} \left(\widetilde{A}(k\omega_0)|\operatorname{sinc}(\omega_\ell-k\omega_0)| - \eta_{k\ell}\right)A_\ell
\end{align}

Using the KKT conditions~\cite{boyd_vandenberghe_2004}, we know for $p(\omega_0, \boldsymbol{\eta}^*)$ to be the maximizer of \eqref{eq:lagrange-rewrite}, the second term should vanish. Given $A_\ell > 0$, we should have that:
\begin{equation}
    \eta_{k\ell} = \widetilde{A}(k\omega_0)|\operatorname{sinc}(\omega_\ell-k\omega_0)|.
\end{equation}
But that means $\rho_f(\omega_0) = p(\omega_0, \boldsymbol{\eta}^*)$ is the exact solution to \eqref{eq:opt2}, \ie, the equality constraint holds perfectly.

Having established that the second optimization problem is a constrained version of the first, it follows that $\Omega$, the feasibility set of \eqref{eq:opt1} is a subset of $\Omega_f$, the feasibility set of \eqref{eq:opt2}. Then, unless $\mathcal{L} = \emptyset$ (which trivially results in $\Omega = \Omega_f$), there exists $\omega_0 \in \Omega\setminus \Omega_f$ such that $\omega_0$ is a valid estimated pitch that has been filtered out by the tube. Therefore, we have shown that the tube will cause shifts in the estimated pitch.

\section{Further Analysis of \name}
\label{app:more_results}

\ifpaper
\else
\begin{figure*}
  \centering
  \scalebox{1}{
       \begin{subfigure}[b]{0.32\textwidth}
         \centering
         \includegraphics[width=\textwidth]{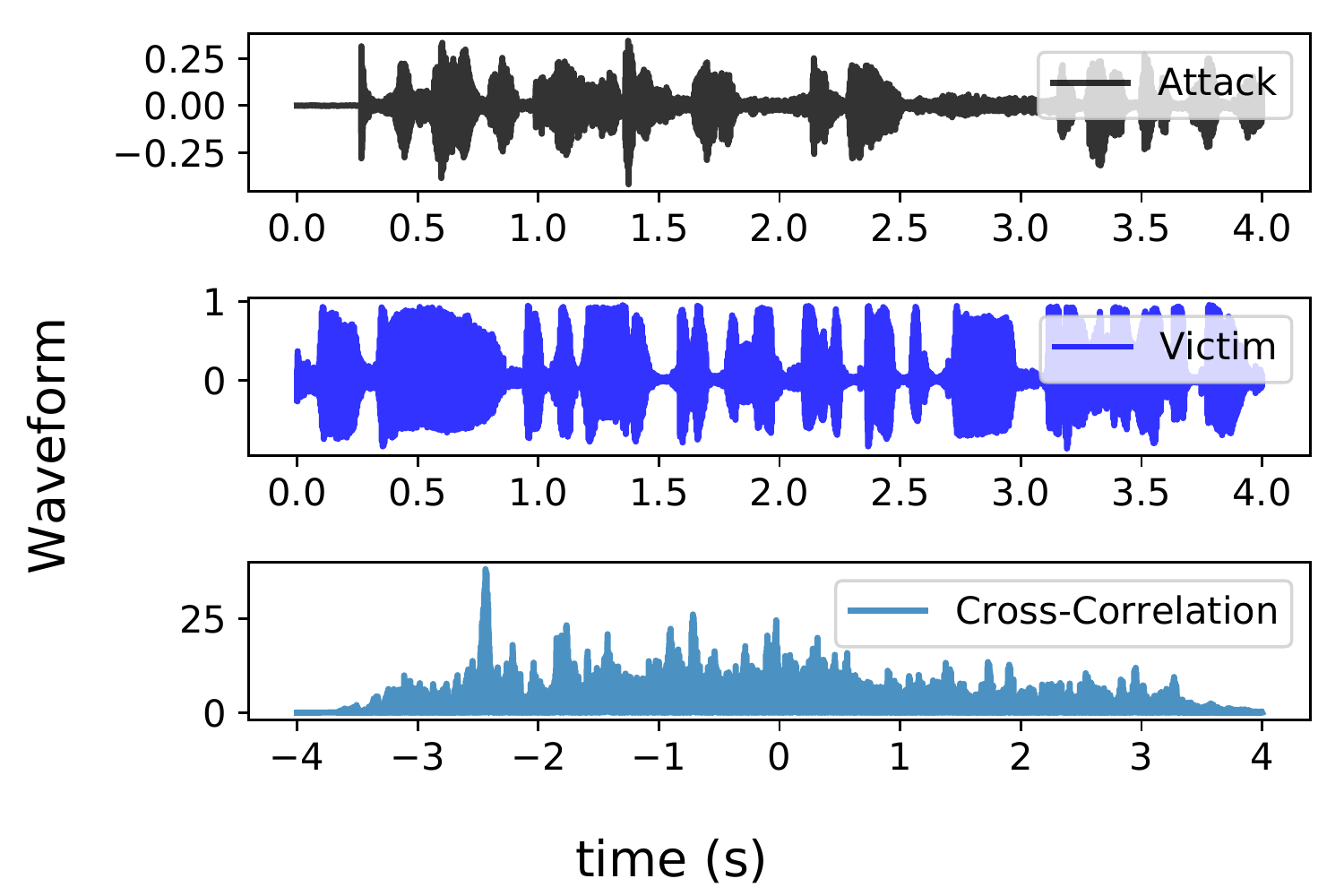}
         \caption{Waveform}
         \label{speech_waveform1}
     \end{subfigure}
     \begin{subfigure}[b]{0.32\textwidth}
         \centering
         \includegraphics[width=\textwidth]{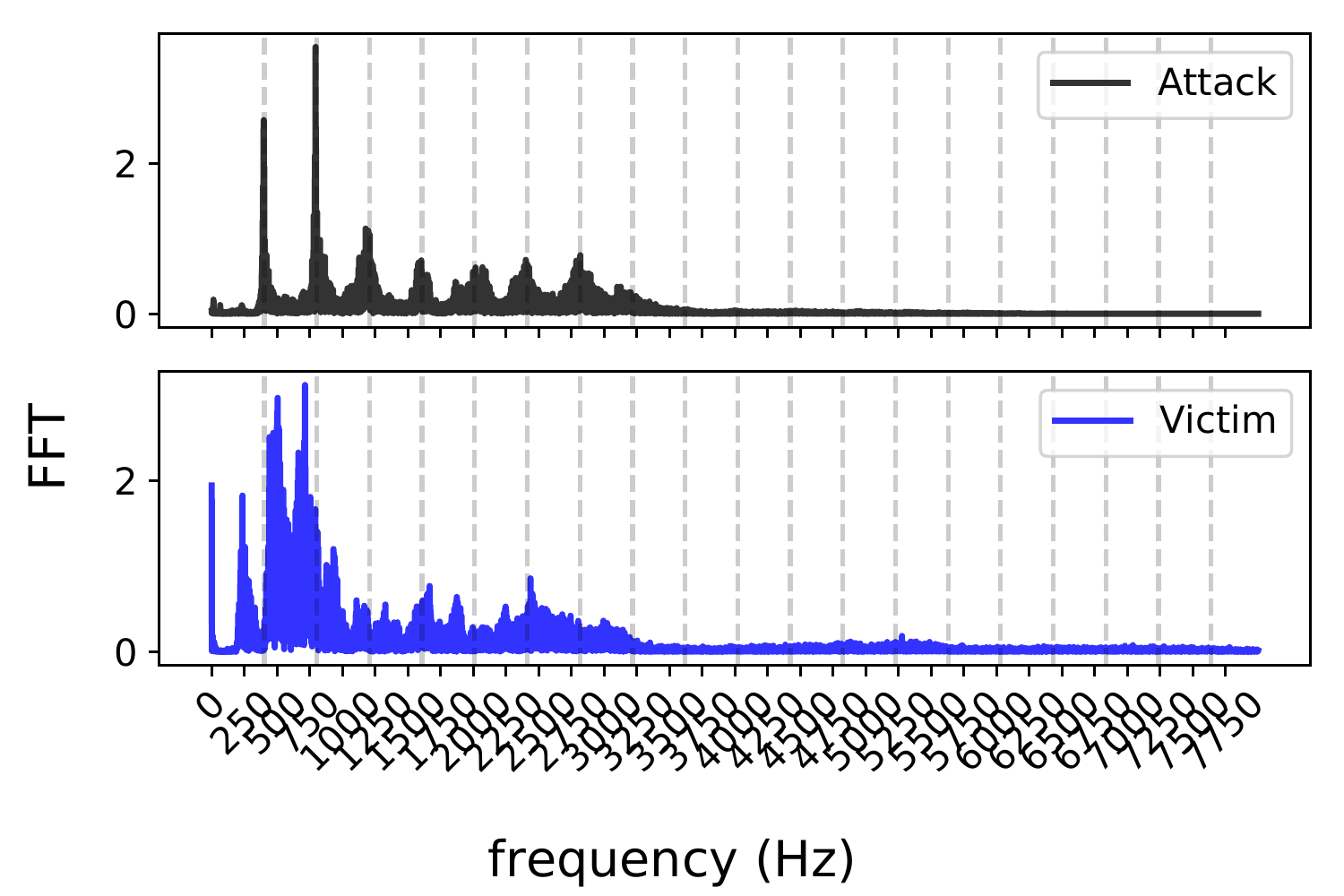}
         \caption{FFT}
         \label{fft_speech1}
     \end{subfigure}
     \begin{subfigure}[b]{0.32\textwidth}
         \centering
         \includegraphics[width=\textwidth]{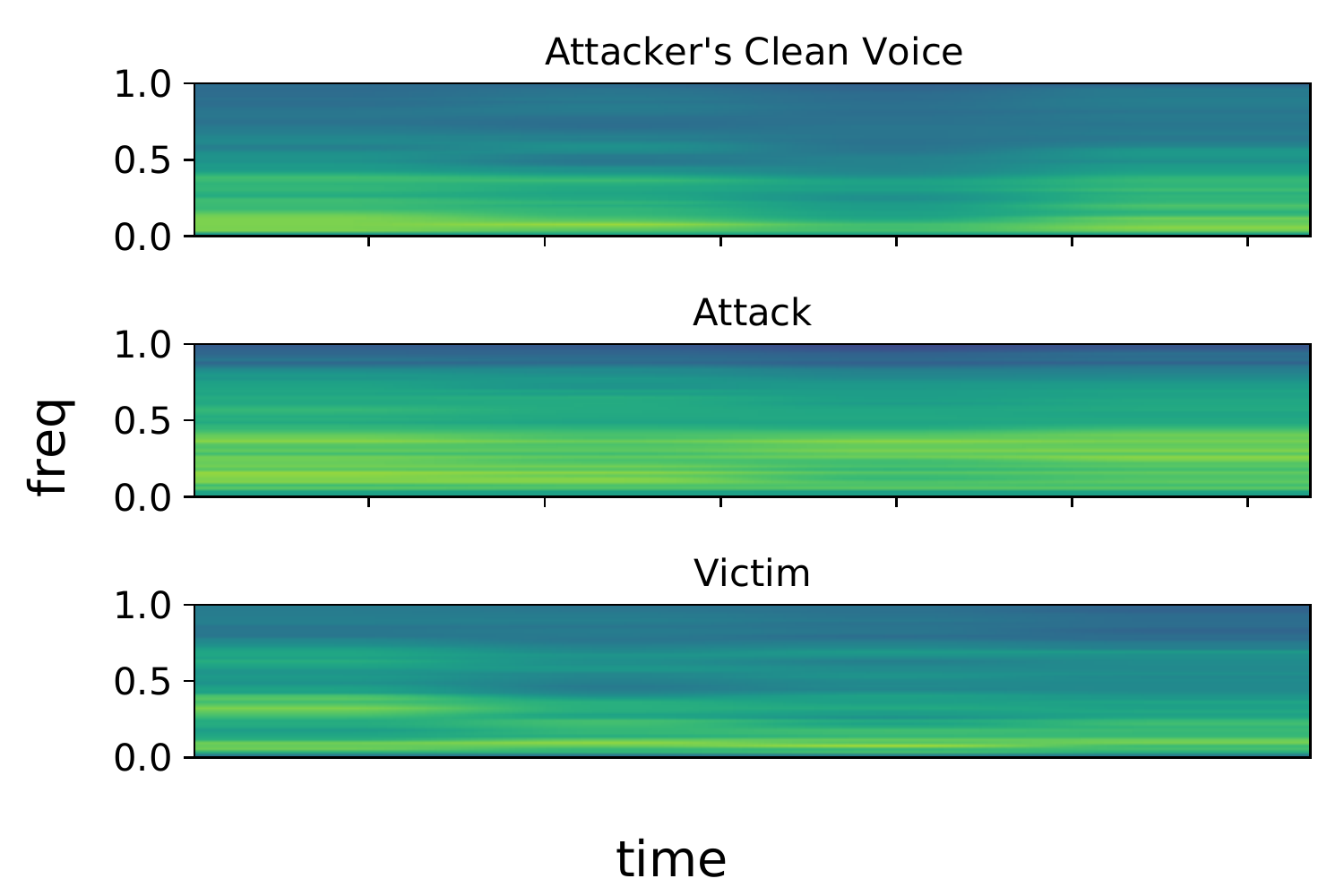}
         \caption{Spectrogram}
         \label{speech_spectrogram1}
     \end{subfigure}
     }
     \caption{\mr{Attack-victim pairs visualization when tube 1 ($L=40.6, d=3.45$ cm) is used: (a) the waveforms and their cross correlation, (b) FFT, and (c) spectrogram for a deeper look at the spectral content. along with the FFT of the BPF model applied to the chirp signal.} }
     \label{fig:attack-victim-signals}
\end{figure*}
\fi

\begin{figure}[ht]
    \centering
    \begin{subfigure}[b]{0.45\columnwidth}
    \centering
    \includegraphics[width=\linewidth]{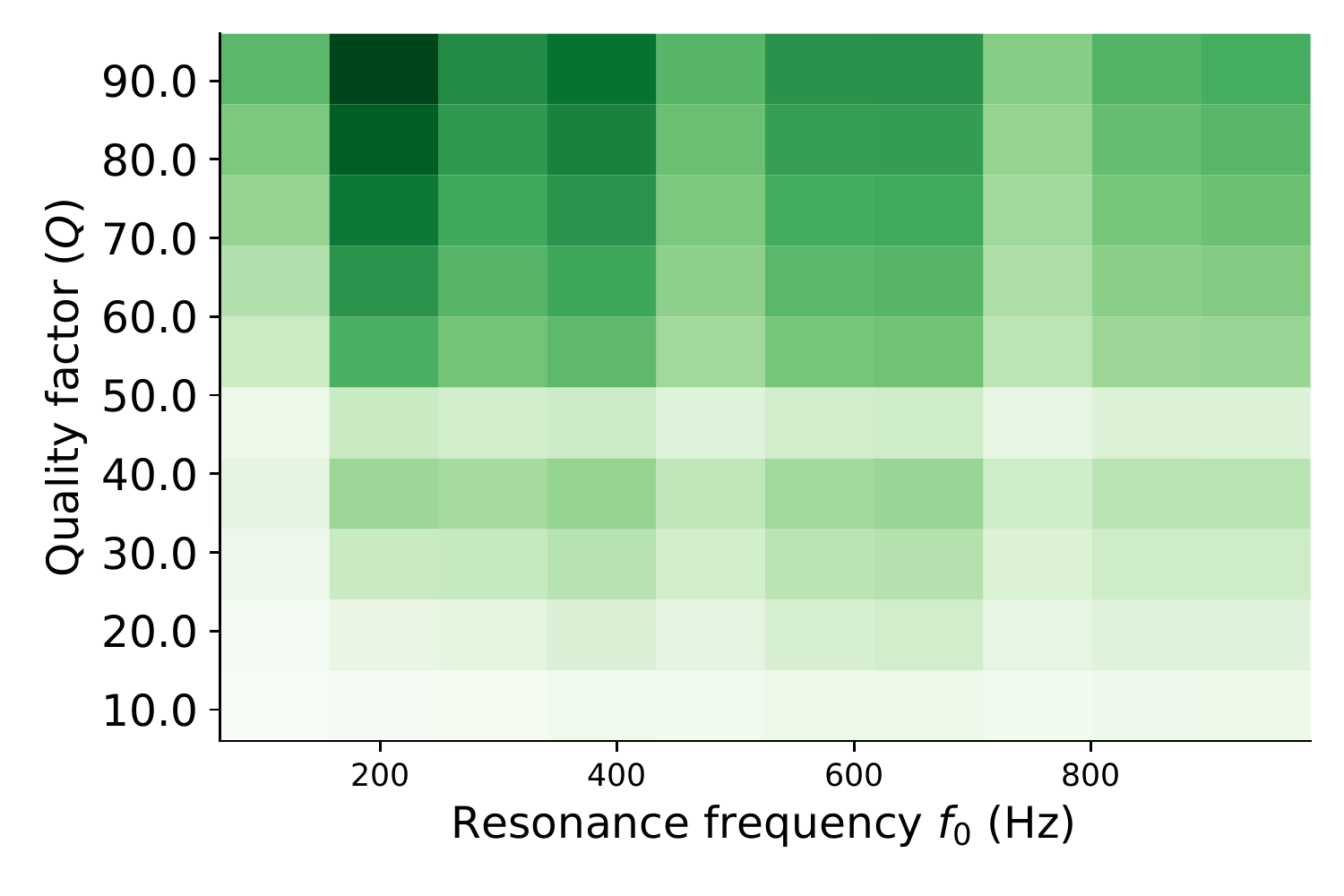}
    \caption{X-Vector}
    \end{subfigure} 
    \begin{subfigure}[b]{0.45\columnwidth}
    \centering
  \includegraphics[width=\linewidth]{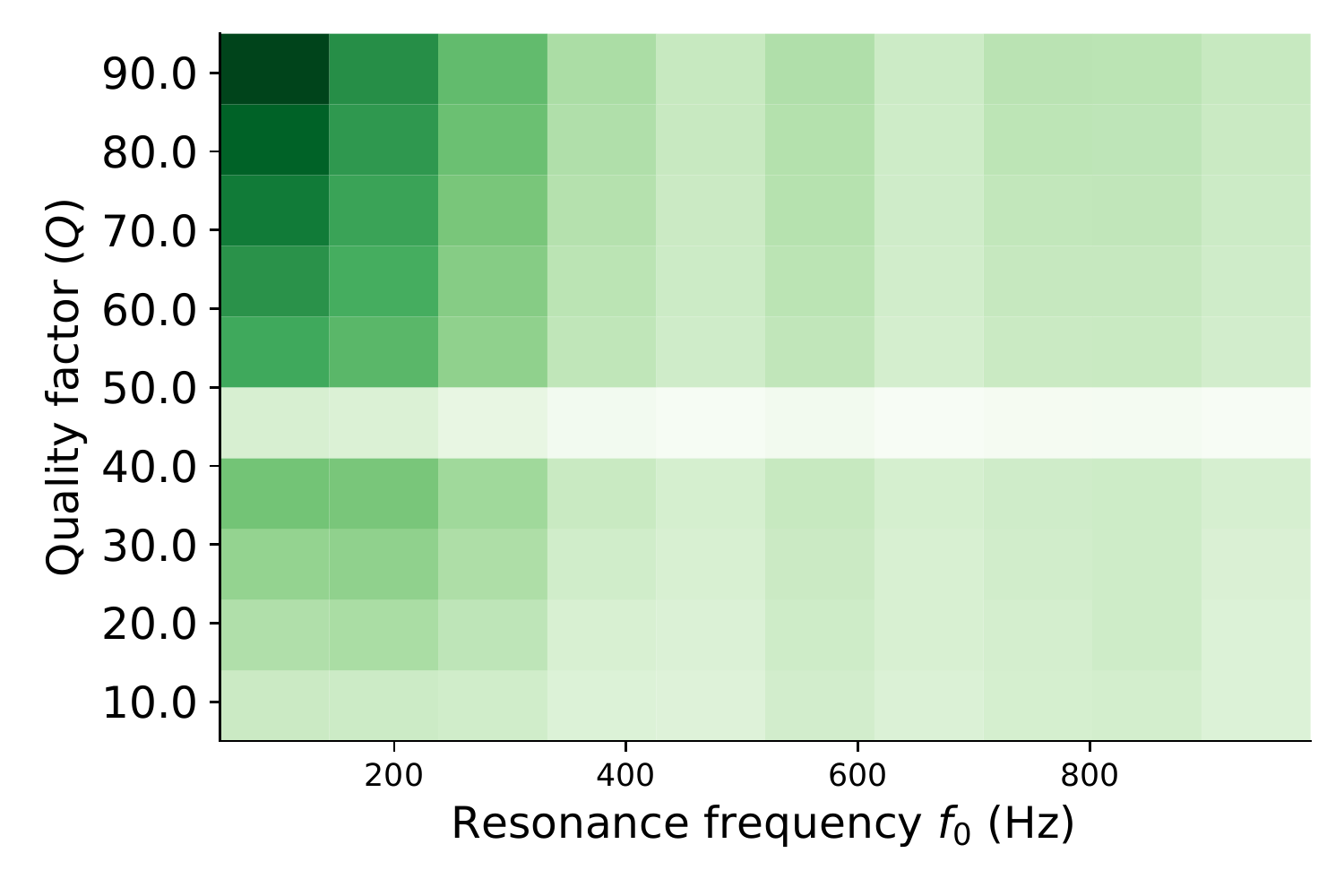}
    \caption{SpeechBrain}
    \end{subfigure} 
    \caption{\mr{Successful impersonations histogram using a single-tube configuration on (a) x-vecotr and (b) SpeechBrain. Most of them are generated by tubes that have Low $f_0$ and high $Q_0$ values.}}
    \label{fig:fQ_histogram}
\end{figure}

\begin{figure}[ht]
    \centering
    \begin{subfigure}[b]{0.45\columnwidth}
    \centering
    \includegraphics[width=\columnwidth]{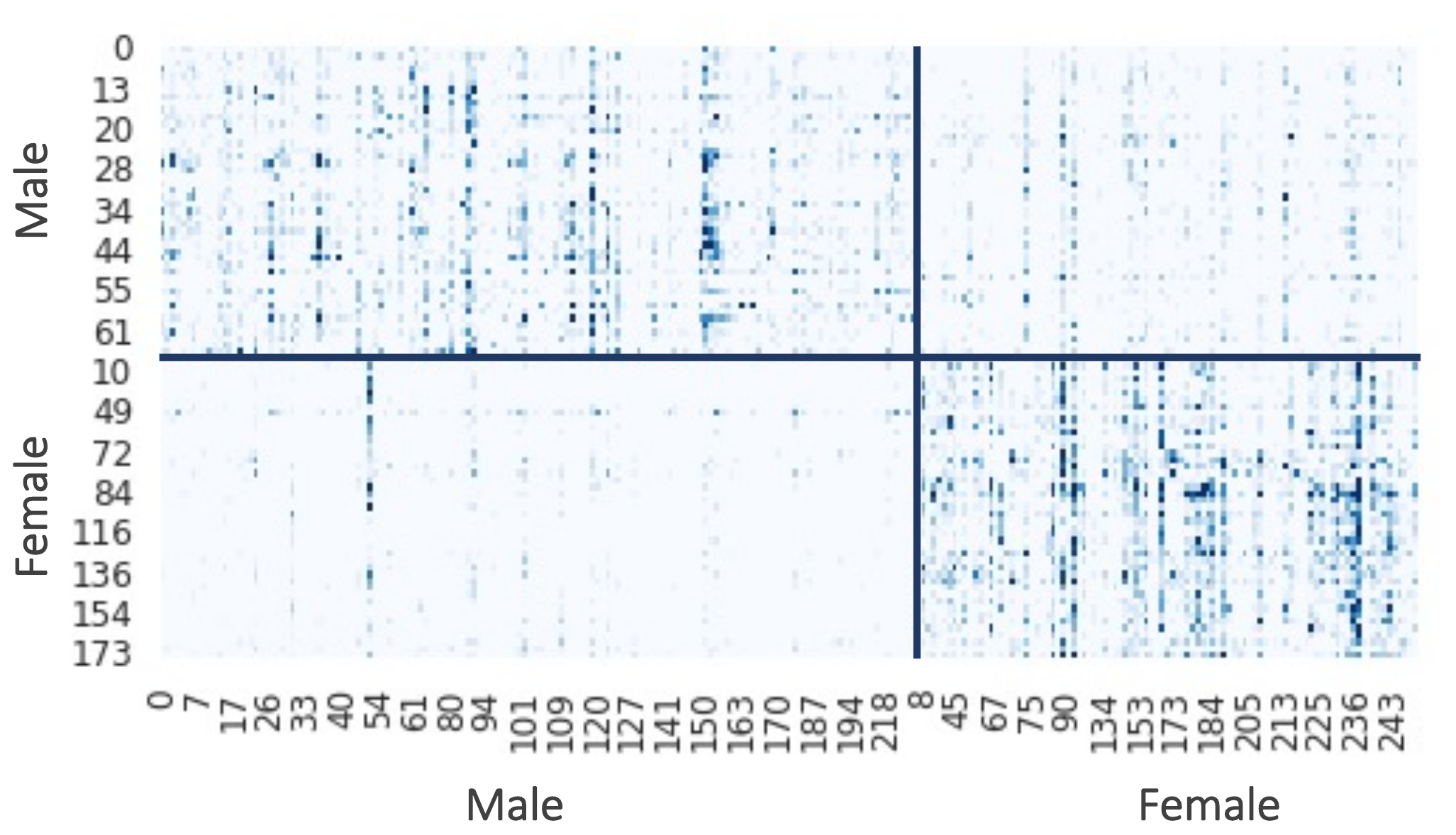}
    \caption{X-Vector}
    \end{subfigure} 
    \begin{subfigure}[b]{0.5\columnwidth}
    \centering
  \includegraphics[width=\columnwidth]{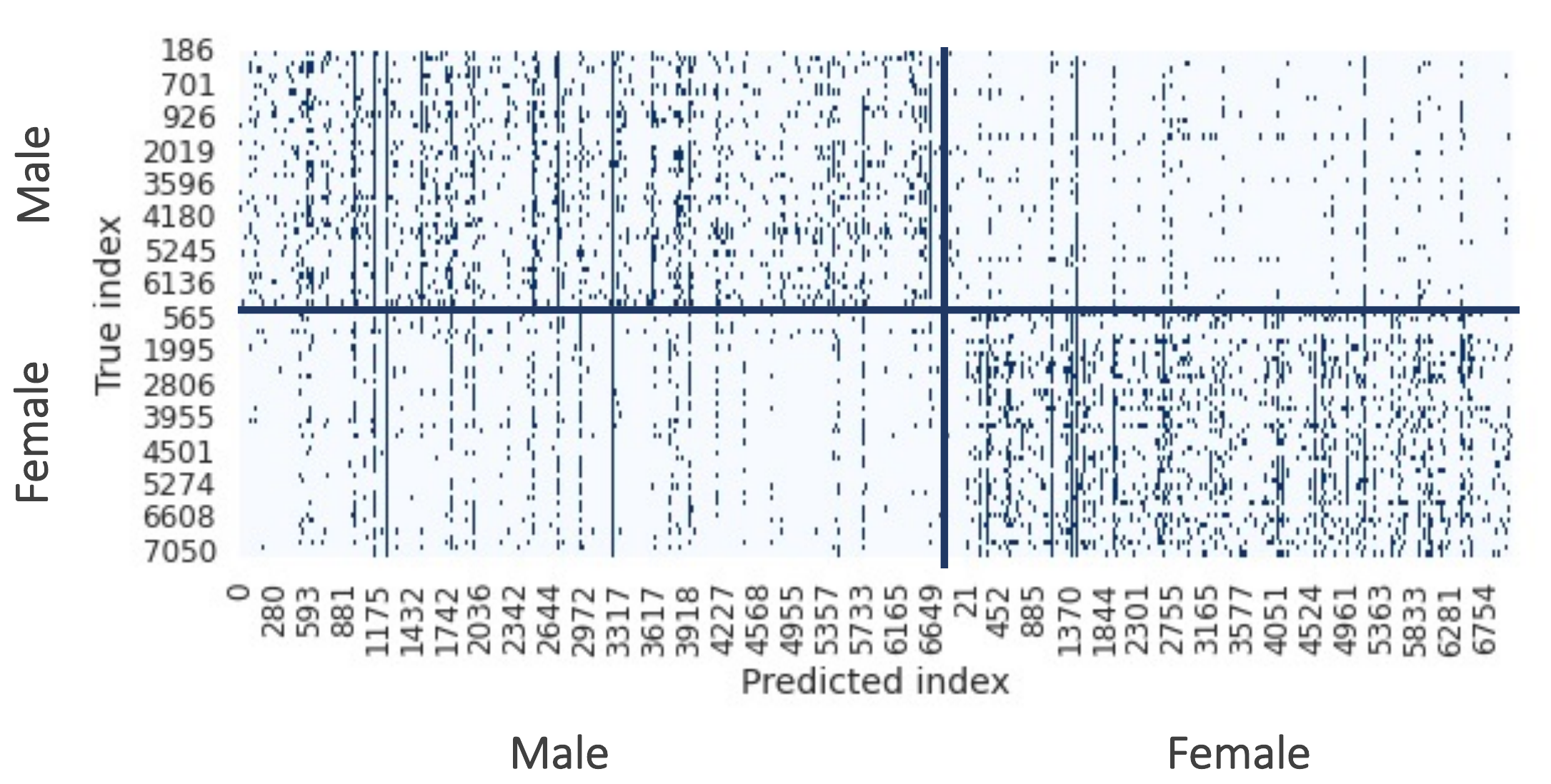}
    \caption{SpeechBrain}
    \end{subfigure} 
    \caption{\mr{The confusion matrix of (a) x-vector and (b) SpeechBrain's predictions on \name attack split by the true (attacker) and predicted (impersonated) speakers sex. The cross-sex submatrix is sparse, indicating attack is more successful within same-sex speakers.}}
    \label{fig:CM_gender}
\end{figure}

\begin{figure}
    \centering
    \scalebox{0.8}{
  \includegraphics[width=\columnwidth]{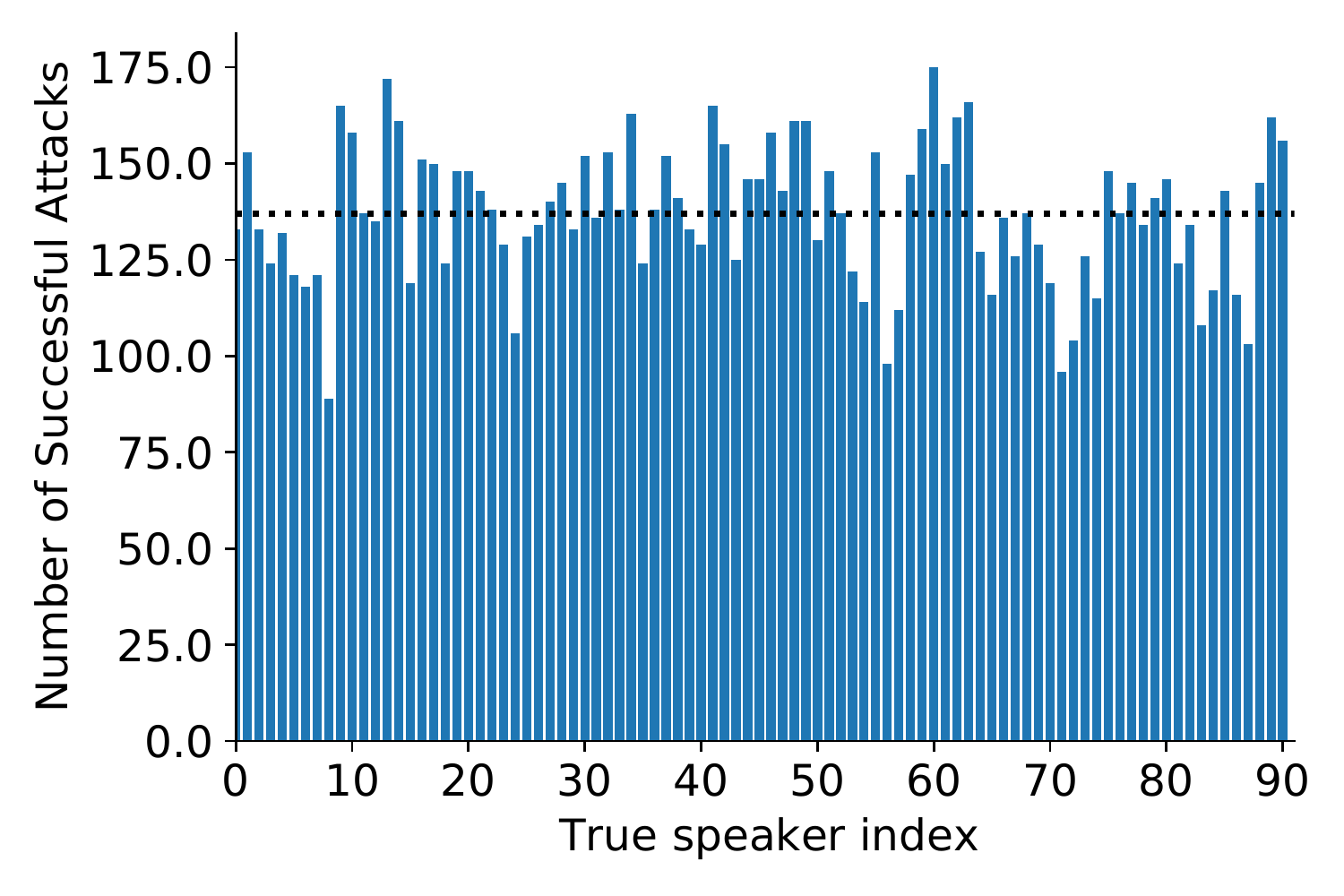}}
    \caption{\mr{Number of successful impersonation attacks (out of 250) on x-vector model  for each adversarial speaker from our VoxCeleb test set.}}
    \label{fig:2tube_target_ids_xv}
\end{figure}

\begin{figure}
    \centering
    \scalebox{0.8}{
  \includegraphics[width=\columnwidth]{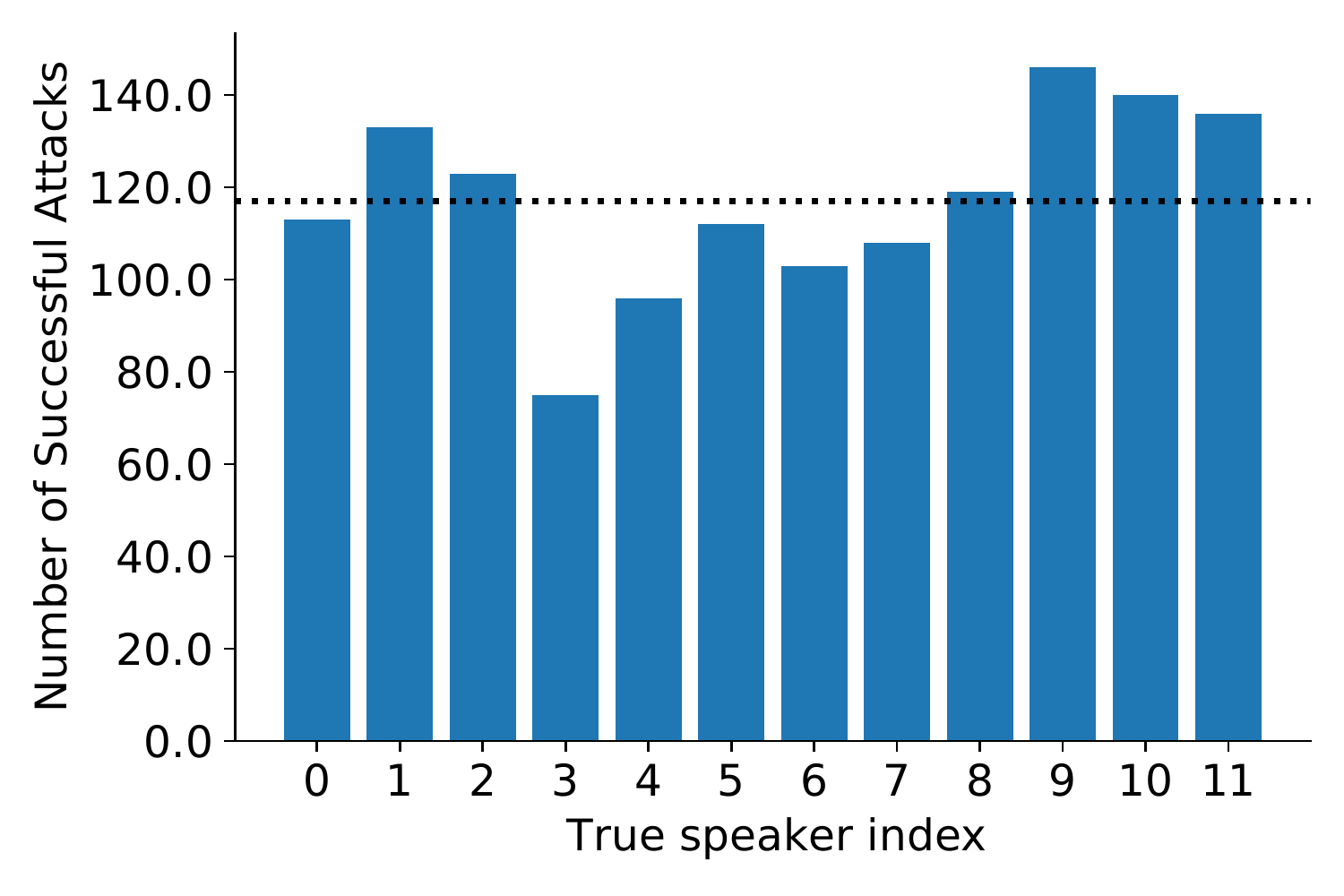}}
    \caption{Number of successful attacks (false predictions) of the x-vector \ASI model on the user study participants recordings.}
    \label{fig:1tube_target_ids_xv_userstudy}
\end{figure}

\begin{table}[ht]
\small
    \centering
\adjustbox{max width=\columnwidth}{%
    \begin{tabular}{l cccccc c}
    \toprule
     \multirow{2}{*}{\textbf{Model}} & \multicolumn{6}{c}{\textbf{Tubes}}\\
     & 1 & 2&3&4&5&6 & Avg\\
    \midrule
    \textbf{X-vector} & 87.53 & 85.8 & 82.45 & 76.47 & 84.22 & 85.95 & 83.74\\
    \textbf{SpeechBrain} & 88.88 & 84.31 & 83.69  & 82.56 2 & 80.38 & 85 & 84.14 \\
    \bottomrule
    \end{tabular}
}
    \caption{\mr{\name's over-the-air predictions consistency rate (\%) across six repeated measurements.}}
    \label{tab:consistency}
\end{table}

\begin{figure}[ht]
  \centering
     \begin{subfigure}[b]{0.45\columnwidth}
         \centering
         \includegraphics[width=\columnwidth]{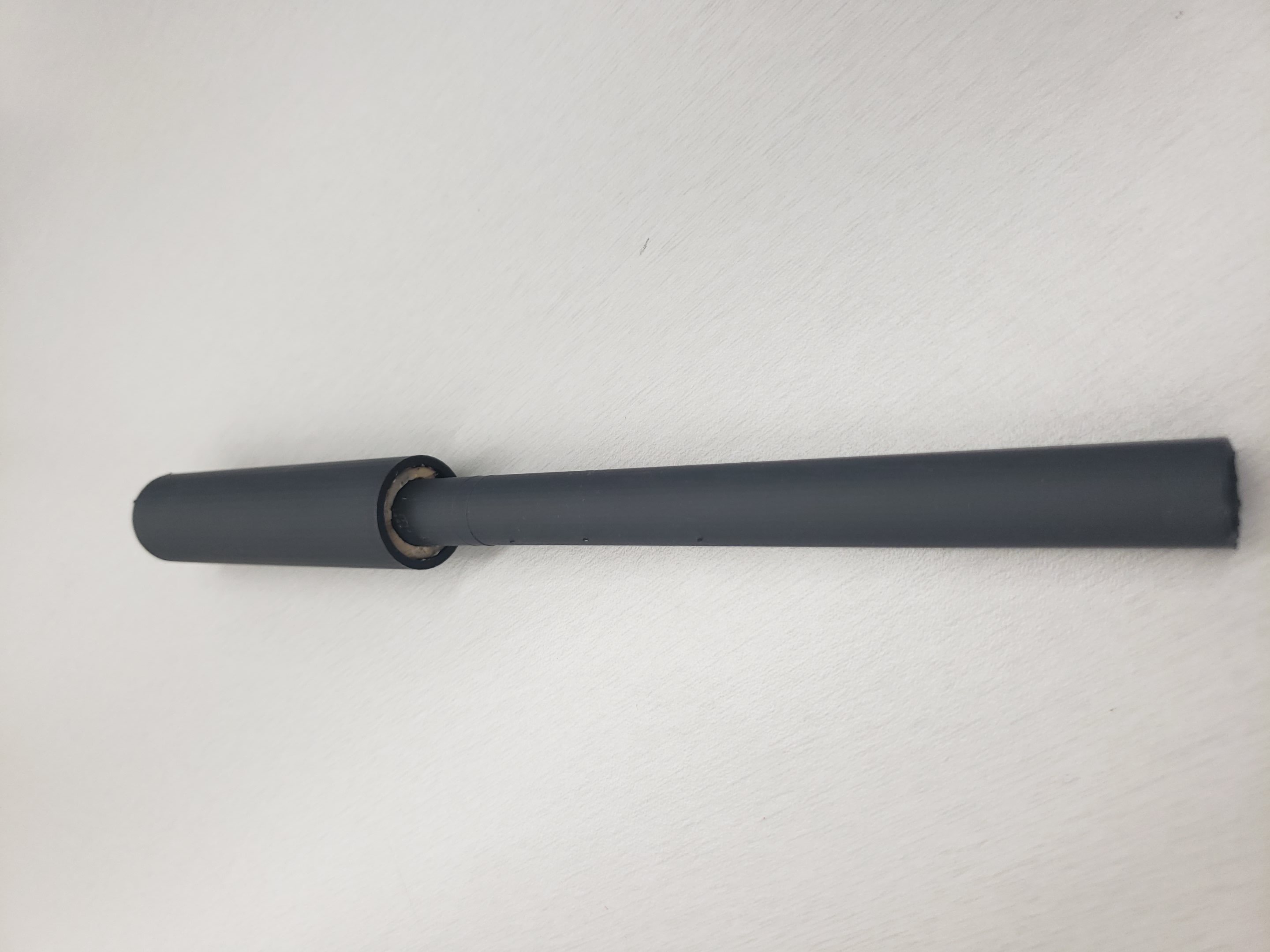}
         \caption{Two tubes connected with a HDF ring}
         \label{2tube_strcuture}
     \end{subfigure}
     \begin{subfigure}[b]{0.45\columnwidth}
         \centering
         \includegraphics[width=\columnwidth]{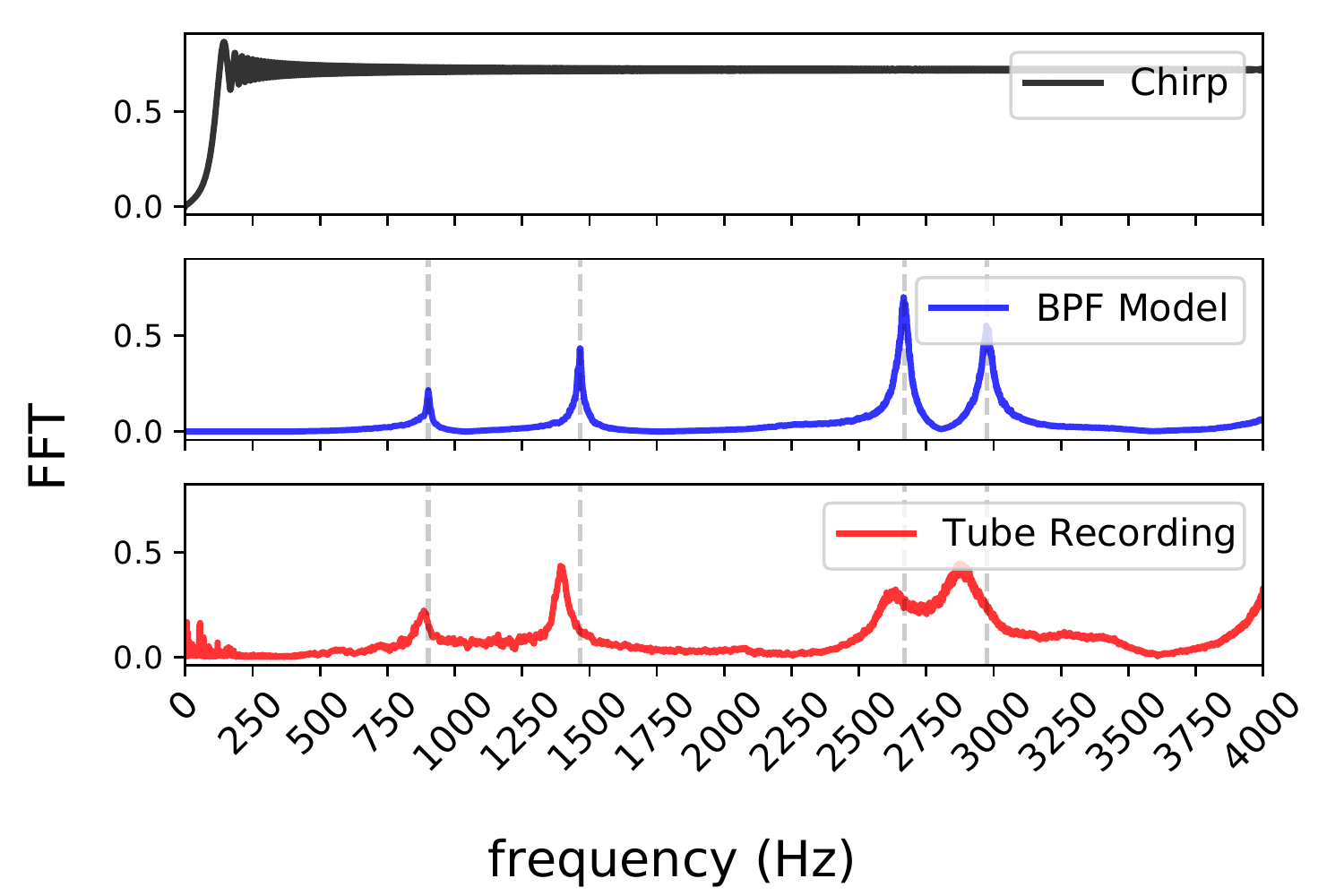}
         \caption{FFT of a chirp}
         \label{2tube_resonance}
     \end{subfigure}
     
     \caption{Two-Tube structure and resonance effect.} 
     \label{fig:2tubes_structure}
\end{figure}

\end{document}